%% file: main.tex
\documentclass[10pt,twocolumn,letterpaper]{article}

\usepackage[pagenumbers]{cvpr} 

\input{preamble}

\definecolor{cvprblue}{rgb}{0.21,0.49,0.74}
\usepackage[pagebackref,breaklinks,colorlinks,allcolors=cvprblue]{hyperref}

\def\confName{CVPR}
\def\confYear{2027}

\title{VersionGS: Compact and Motion-Robust Dynamic Gaussian Splatting via Versioned Gaussians}

\author{Shihang Wei \qquad Yichi Zhang \qquad Fengqing Zhu\\
Purdue University\\
{\tt\small \{shihang.wei, zhan5096, zhu0\}@purdue.edu}
}

\makeatletter\patchcmd{\@maketitle}{\vskip .375in}{\vskip .1in}{}{\errmessage{title patch A}}\patchcmd{\@maketitle}{\vspace*{24pt}}{\vspace*{12pt}}{}{\errmessage{title patch B}}\makeatother
\begin{document}
\twocolumn[{%
\renewcommand\twocolumn[1][]{#1}%
\maketitle
\vspace{-12pt}
\begin{center}
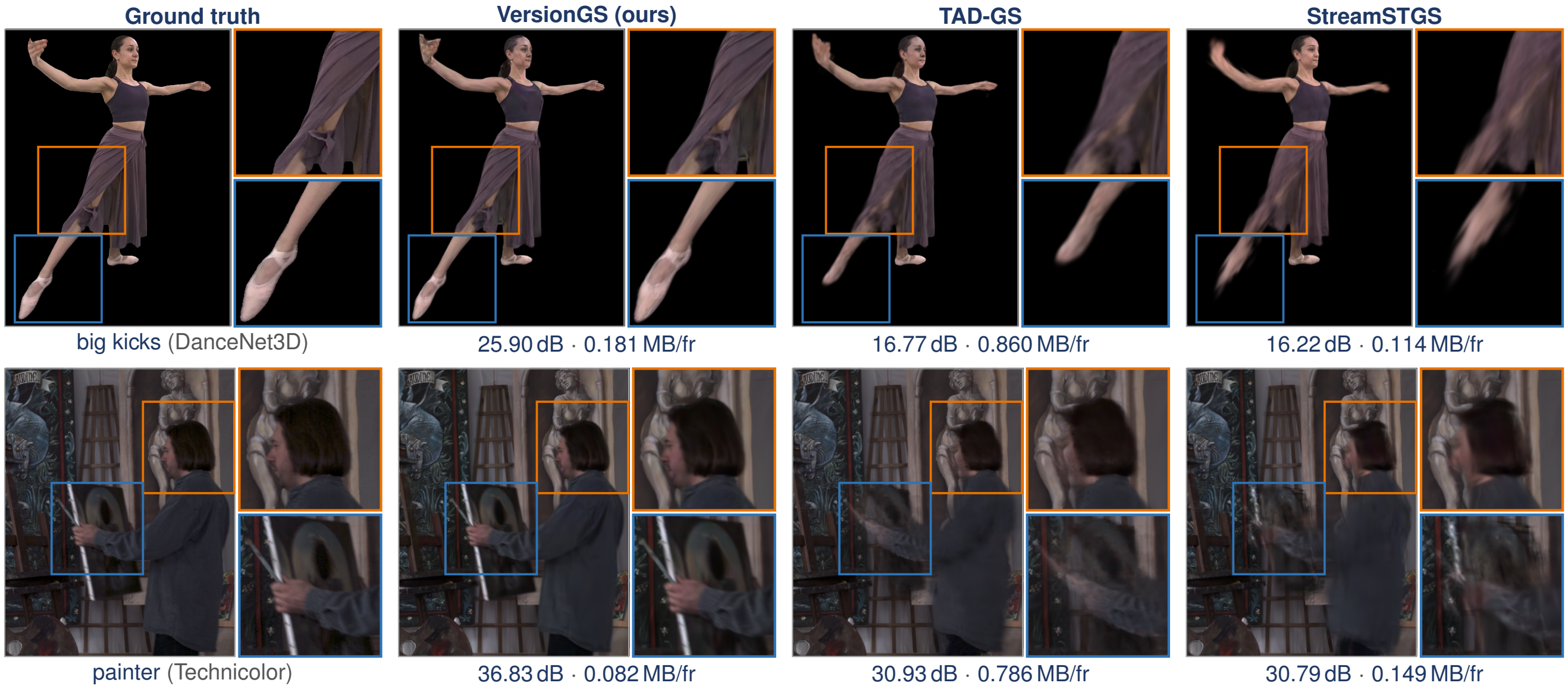

  \captionsetup{type=figure}
  \input{fig/teaser}
  \captionof{figure}{Held-out views of a fast dance, big kicks from DanceNet3D, and of a painter moving a canvas from Technicolor. TAD-GS and StreamSTGS smear the skirt, lose the ballet shoe, blur the painter's face and dissolve the canvas. VersionGS achieves best quality with low rate. The numbers give the PSNR, in the person mask for big kicks, and each method's average storage per frame, ours at $\rho=0.35$.}
  \label{fig:teaser}
\end{center}
\vspace{-1mm}
}]
\input{sec/0_abstract}

\input{sec/1_intro}
\input{sec/2_related}
\input{sec/3_method}
\input{sec/4_experiments}

\input{sec/llm_statement}
{
    \small
    \bibliographystyle{ieeenat_fullname}
    \bibliography{main}
}

\input{sec/X_suppl}

\end{document}

%% file: preamble.tex
\newcommand{\baselineKeys}{lee2024ex4dgs,yang2023gs4d,sun20243dgstream,zhang2025dfcgs,ke2026streamstgs,girish2024queen,sandu2026tadgs}
\newcommand{\citeBaselines}{\expandafter\cite\expandafter{\baselineKeys}}
\usepackage{tikz}
\usetikzlibrary{shapes.geometric,arrows.meta}
\usepackage{pgfplots}
\pgfplotsset{compat=1.18}
\usepackage{colortbl}
\usepackage{multirow}   %
\newcommand{\rankA}[1]{\cellcolor{red!25}#1}
\newcommand{\rankB}[1]{\cellcolor{orange!25}#1}
\newcommand{\rankC}[1]{\cellcolor{yellow!35}#1}

\usepackage{microtype}

\renewcommand{\paragraph}[1]{\vspace{.2em}\noindent\textbf{#1}}



%% file: fig/teaser.tex
\input{fig/theme}\begin{tikzpicture}[font=\sffamily, x=1cm, y=1cm]
  \useasboundingbox (-0.02,-0.02) rectangle (17.37,7.62);
  \node[tzlabel, font=\sffamily\bfseries\scriptsize] at (2.115,7.50) {Ground truth};
  \node[tzlabel, font=\sffamily\bfseries\scriptsize] at (6.495,7.50) {VersionGS (ours)};
  \node[tzlabel, font=\sffamily\bfseries\scriptsize] at (10.875,7.50) {TAD-GS};
  \node[tzlabel, font=\sffamily\bfseries\scriptsize] at (15.255,7.50) {StreamSTGS};
  \node[anchor=south west, inner sep=0pt] at (0.024,4.050) {\includegraphics[width=2.507cm,height=3.300cm]{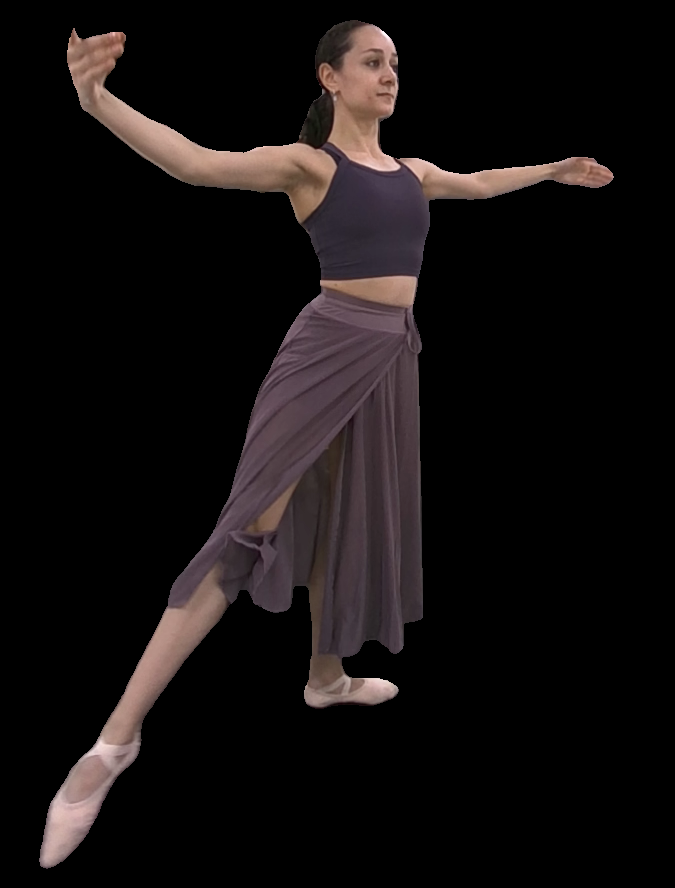}};
  \draw[tzAxisLine, line width=.4pt] (0.024,4.050) rectangle ++(2.507,3.300);
  \draw[tzBorn, line width=.7pt] (0.395,5.079) rectangle ++(0.966,0.966);
  \node[anchor=south west, inner sep=0pt] at (2.581,5.725) {\includegraphics[width=1.625cm,height=1.625cm]{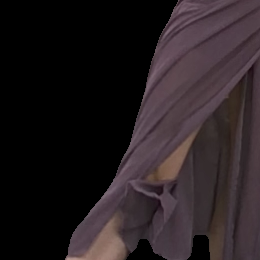}};
  \draw[tzBorn, line width=.8pt] (2.581,5.725) rectangle ++(1.625,1.625);
  \draw[tzBlue, line width=.7pt] (0.135,4.095) rectangle ++(0.966,0.966);
  \node[anchor=south west, inner sep=0pt] at (2.581,4.050) {\includegraphics[width=1.625cm,height=1.625cm]{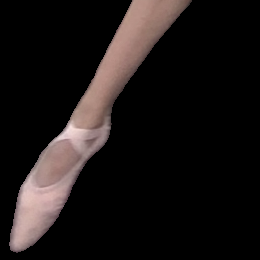}};
  \draw[tzBlue, line width=.8pt] (2.581,4.050) rectangle ++(1.625,1.625);
  \node[tzlabel] at (2.115,3.860) {big kicks \textcolor{tzAxis}{(DanceNet3D)}};
  \node[anchor=south west, inner sep=0pt] at (4.404,4.050) {\includegraphics[width=2.507cm,height=3.300cm]{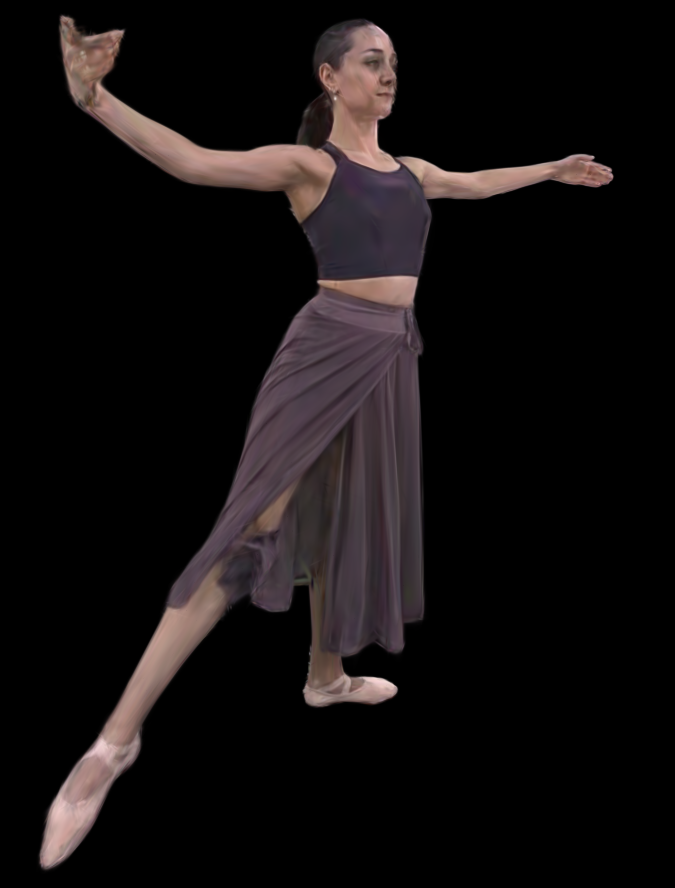}};
  \draw[tzAxisLine, line width=.4pt] (4.404,4.050) rectangle ++(2.507,3.300);
  \draw[tzBorn, line width=.7pt] (4.775,5.079) rectangle ++(0.966,0.966);
  \node[anchor=south west, inner sep=0pt] at (6.961,5.725) {\includegraphics[width=1.625cm,height=1.625cm]{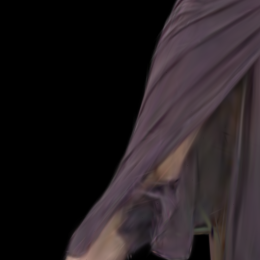}};
  \draw[tzBorn, line width=.8pt] (6.961,5.725) rectangle ++(1.625,1.625);
  \draw[tzBlue, line width=.7pt] (4.515,4.095) rectangle ++(0.966,0.966);
  \node[anchor=south west, inner sep=0pt] at (6.961,4.050) {\includegraphics[width=1.625cm,height=1.625cm]{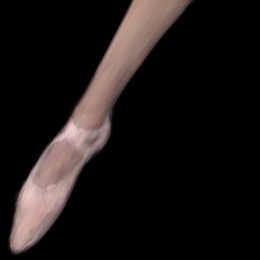}};
  \draw[tzBlue, line width=.8pt] (6.961,4.050) rectangle ++(1.625,1.625);
  \node[tzlabel] at (6.495,3.860) {25.90\,dB \textcolor{tzAxis}{$\cdot$} 0.181\,MB/fr};
  \node[anchor=south west, inner sep=0pt] at (8.784,4.050) {\includegraphics[width=2.507cm,height=3.300cm]{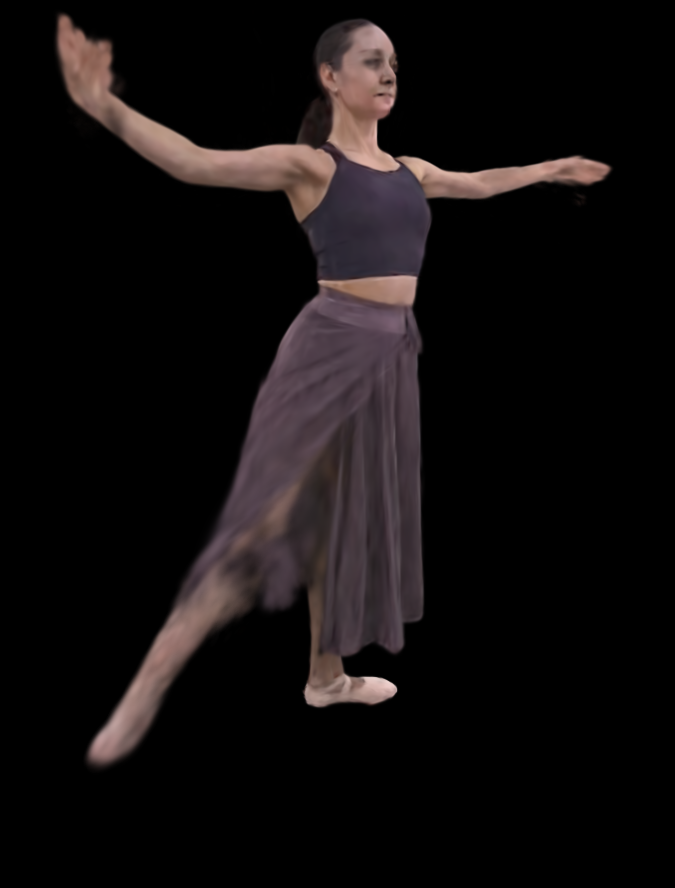}};
  \draw[tzAxisLine, line width=.4pt] (8.784,4.050) rectangle ++(2.507,3.300);
  \draw[tzBorn, line width=.7pt] (9.155,5.079) rectangle ++(0.966,0.966);
  \node[anchor=south west, inner sep=0pt] at (11.341,5.725) {\includegraphics[width=1.625cm,height=1.625cm]{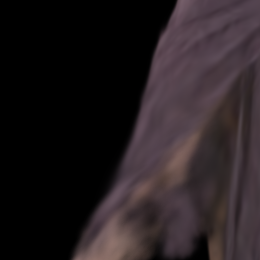}};
  \draw[tzBorn, line width=.8pt] (11.341,5.725) rectangle ++(1.625,1.625);
  \draw[tzBlue, line width=.7pt] (8.895,4.095) rectangle ++(0.966,0.966);
  \node[anchor=south west, inner sep=0pt] at (11.341,4.050) {\includegraphics[width=1.625cm,height=1.625cm]{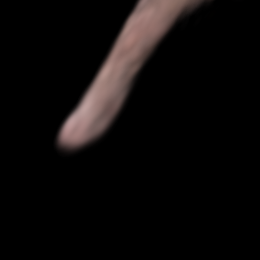}};
  \draw[tzBlue, line width=.8pt] (11.341,4.050) rectangle ++(1.625,1.625);
  \node[tzlabel] at (10.875,3.860) {16.77\,dB \textcolor{tzAxis}{$\cdot$} 0.860\,MB/fr};
  \node[anchor=south west, inner sep=0pt] at (13.164,4.050) {\includegraphics[width=2.507cm,height=3.300cm]{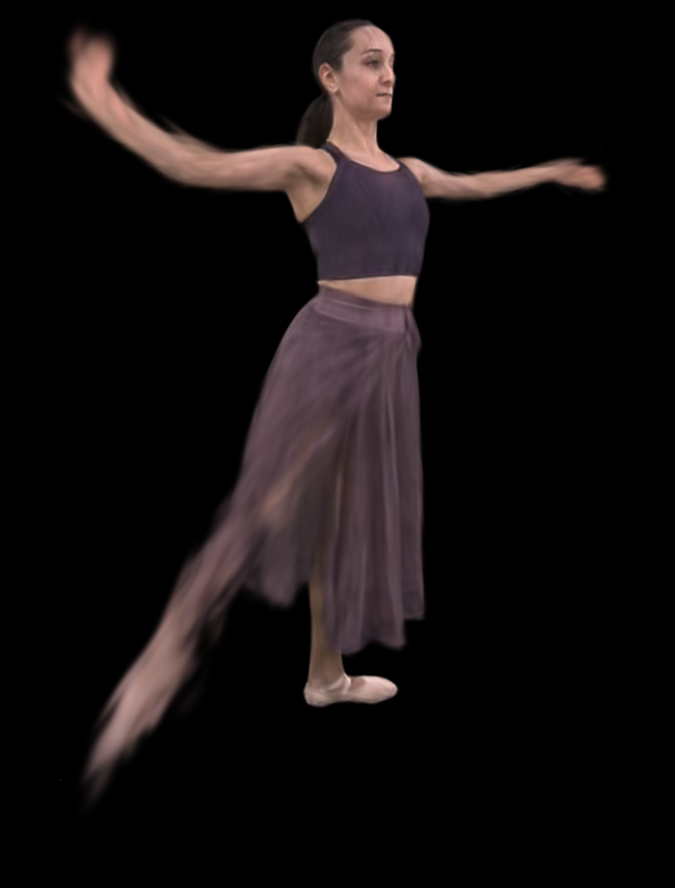}};
  \draw[tzAxisLine, line width=.4pt] (13.164,4.050) rectangle ++(2.507,3.300);
  \draw[tzBorn, line width=.7pt] (13.535,5.079) rectangle ++(0.966,0.966);
  \node[anchor=south west, inner sep=0pt] at (15.721,5.725) {\includegraphics[width=1.625cm,height=1.625cm]{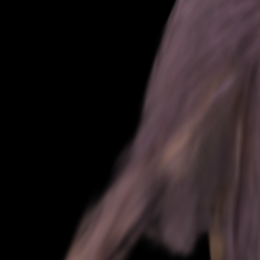}};
  \draw[tzBorn, line width=.8pt] (15.721,5.725) rectangle ++(1.625,1.625);
  \draw[tzBlue, line width=.7pt] (13.275,4.095) rectangle ++(0.966,0.966);
  \node[anchor=south west, inner sep=0pt] at (15.721,4.050) {\includegraphics[width=1.625cm,height=1.625cm]{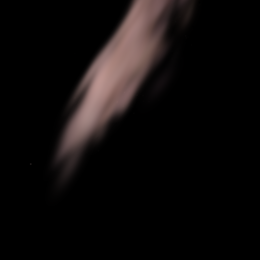}};
  \draw[tzBlue, line width=.8pt] (15.721,4.050) rectangle ++(1.625,1.625);
  \node[tzlabel] at (15.255,3.860) {16.22\,dB \textcolor{tzAxis}{$\cdot$} 0.114\,MB/fr};
  \node[anchor=south west, inner sep=0pt] at (0.023,0.380) {\includegraphics[width=2.559cm,height=3.200cm]{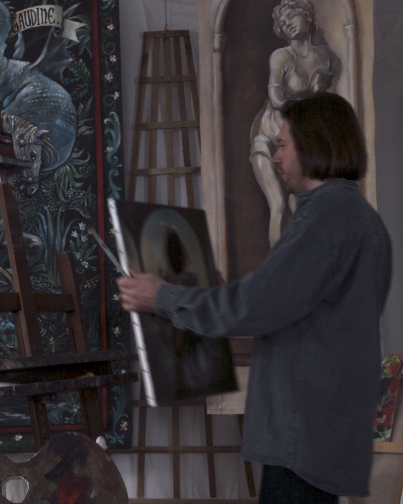}};
  \draw[tzAxisLine, line width=.4pt] (0.023,0.380) rectangle ++(2.559,3.200);
  \draw[tzBorn, line width=.7pt] (1.560,2.196) rectangle ++(1.016,1.016);
  \node[anchor=south west, inner sep=0pt] at (2.632,2.005) {\includegraphics[width=1.575cm,height=1.575cm]{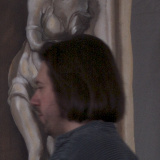}};
  \draw[tzBorn, line width=.8pt] (2.632,2.005) rectangle ++(1.575,1.575);
  \draw[tzBlue, line width=.7pt] (0.544,1.294) rectangle ++(1.016,1.016);
  \node[anchor=south west, inner sep=0pt] at (2.632,0.380) {\includegraphics[width=1.575cm,height=1.575cm]{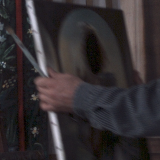}};
  \draw[tzBlue, line width=.8pt] (2.632,0.380) rectangle ++(1.575,1.575);
  \node[tzlabel] at (2.115,0.190) {painter \textcolor{tzAxis}{(Technicolor)}};
  \node[anchor=south west, inner sep=0pt] at (4.403,0.380) {\includegraphics[width=2.559cm,height=3.200cm]{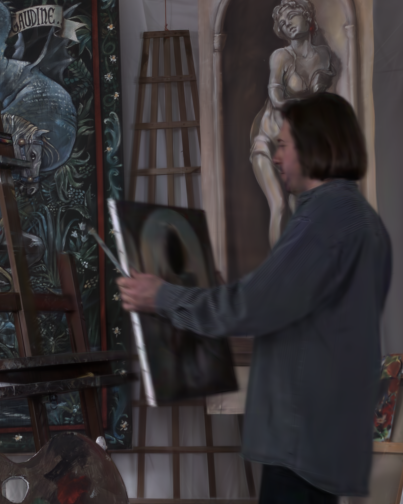}};
  \draw[tzAxisLine, line width=.4pt] (4.403,0.380) rectangle ++(2.559,3.200);
  \draw[tzBorn, line width=.7pt] (5.940,2.196) rectangle ++(1.016,1.016);
  \node[anchor=south west, inner sep=0pt] at (7.012,2.005) {\includegraphics[width=1.575cm,height=1.575cm]{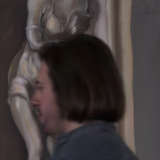}};
  \draw[tzBorn, line width=.8pt] (7.012,2.005) rectangle ++(1.575,1.575);
  \draw[tzBlue, line width=.7pt] (4.924,1.294) rectangle ++(1.016,1.016);
  \node[anchor=south west, inner sep=0pt] at (7.012,0.380) {\includegraphics[width=1.575cm,height=1.575cm]{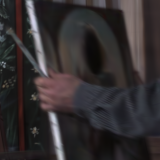}};
  \draw[tzBlue, line width=.8pt] (7.012,0.380) rectangle ++(1.575,1.575);
  \node[tzlabel] at (6.495,0.190) {36.83\,dB \textcolor{tzAxis}{$\cdot$} 0.082\,MB/fr};
  \node[anchor=south west, inner sep=0pt] at (8.783,0.380) {\includegraphics[width=2.559cm,height=3.200cm]{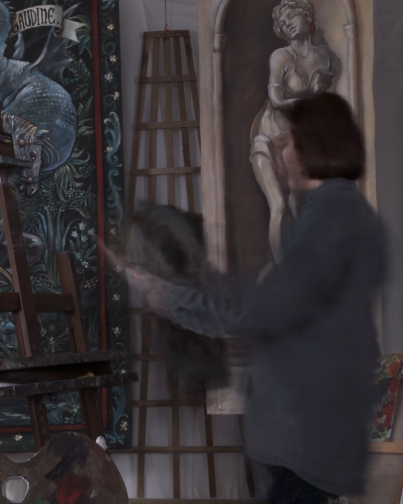}};
  \draw[tzAxisLine, line width=.4pt] (8.783,0.380) rectangle ++(2.559,3.200);
  \draw[tzBorn, line width=.7pt] (10.320,2.196) rectangle ++(1.016,1.016);
  \node[anchor=south west, inner sep=0pt] at (11.392,2.005) {\includegraphics[width=1.575cm,height=1.575cm]{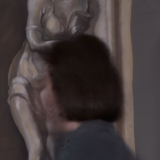}};
  \draw[tzBorn, line width=.8pt] (11.392,2.005) rectangle ++(1.575,1.575);
  \draw[tzBlue, line width=.7pt] (9.304,1.294) rectangle ++(1.016,1.016);
  \node[anchor=south west, inner sep=0pt] at (11.392,0.380) {\includegraphics[width=1.575cm,height=1.575cm]{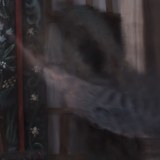}};
  \draw[tzBlue, line width=.8pt] (11.392,0.380) rectangle ++(1.575,1.575);
  \node[tzlabel] at (10.875,0.190) {30.93\,dB \textcolor{tzAxis}{$\cdot$} 0.786\,MB/fr};
  \node[anchor=south west, inner sep=0pt] at (13.163,0.380) {\includegraphics[width=2.559cm,height=3.200cm]{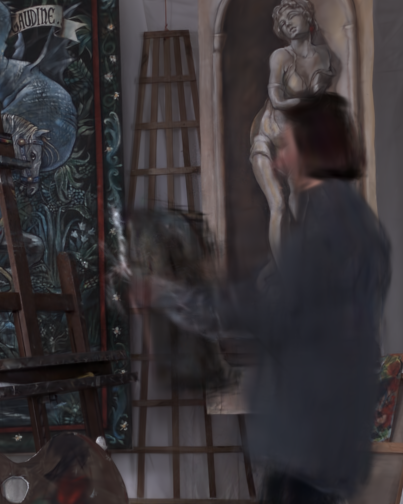}};
  \draw[tzAxisLine, line width=.4pt] (13.163,0.380) rectangle ++(2.559,3.200);
  \draw[tzBorn, line width=.7pt] (14.700,2.196) rectangle ++(1.016,1.016);
  \node[anchor=south west, inner sep=0pt] at (15.772,2.005) {\includegraphics[width=1.575cm,height=1.575cm]{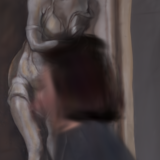}};
  \draw[tzBorn, line width=.8pt] (15.772,2.005) rectangle ++(1.575,1.575);
  \draw[tzBlue, line width=.7pt] (13.684,1.294) rectangle ++(1.016,1.016);
  \node[anchor=south west, inner sep=0pt] at (15.772,0.380) {\includegraphics[width=1.575cm,height=1.575cm]{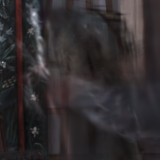}};
  \draw[tzBlue, line width=.8pt] (15.772,0.380) rectangle ++(1.575,1.575);
  \node[tzlabel] at (15.255,0.190) {30.79\,dB \textcolor{tzAxis}{$\cdot$} 0.149\,MB/fr};
\end{tikzpicture}%

%% file: fig/theme.tex
\definecolor{tzNavy}{HTML}{1F3864}%
\definecolor{tzBlue}{HTML}{2E74B5}%
\definecolor{tzBlueMid}{HTML}{5898D4}%
\definecolor{tzBlueLight}{HTML}{DEEBF7}%
\definecolor{tzBoxLine}{HTML}{96B9DA}%
\definecolor{tzPanel}{HTML}{EEF3FA}%
\definecolor{tzPanelLine}{HTML}{C7D5E5}%
\definecolor{tzBrown}{HTML}{8A5A11}%
\definecolor{tzRust}{HTML}{C45810}%
\definecolor{tzReuse}{HTML}{A9A9A9}%
\definecolor{tzNew}{HTML}{2E74B5}%
\definecolor{tzBorn}{HTML}{E97300}%
\definecolor{tzAxis}{HTML}{555555}%
\definecolor{tzAxisLine}{HTML}{9F9F9F}%
\definecolor{tzGrid}{HTML}{E2E2E2}%
\tikzset{%
  tzpanel/.style={draw=tzPanelLine, fill=tzPanel, line width=.8pt, rounded corners=4pt},%
  tztitle/.style={anchor=west, text=tzNavy, font=\sffamily\bfseries\footnotesize, inner sep=0pt},%
  tzmodule/.style={draw=tzBlueMid, fill=white, line width=.9pt, rounded corners=3pt, text=tzNavy, align=center,
    font=\sffamily\scriptsize, inner sep=2pt},%
  tzbox/.style={draw=tzBoxLine, fill=tzBlueLight, line width=.7pt, rounded corners=2pt, text=tzNavy, align=center,
    font=\sffamily\scriptsize, inner sep=2pt},%
  tzbig/.style={-{Triangle[length=1.7mm,width=3.6mm]}, line width=1.5mm, tzNavy},%
  tzarr/.style={-{Stealth[length=1.6mm]}, line width=.7pt, tzNavy},%
  tzlabel/.style={font=\sffamily\scriptsize, text=tzNavy, inner sep=1pt},%
}%
\def\tzG#1#2#3#4#5#6{\fill[#1, rotate around={#6:(#2,#3)}] (#2,#3) ellipse ({#4} and {#5});}%

%% file: sec/0_abstract.tex
\patchcmd{\abstract}{\vspace*{12pt}}{\vspace*{2pt}}{}{\PackageWarning{abstract}{gap patch failed}}%
\begin{abstract}
Recent dynamic Gaussian splatting methods encode temporal change, either as smooth functions of time or as constrained updates between frames.
We observe that their reconstruction quality degrades significantly under fast motion.
In challenging dance sequences, their PSNR drops by 1.1–4.0 dB on high-motion frames compared to low-motion frames.
To overcome this limitation, we propose VersionGS, a method that directly encodes per-frame Gaussian states and uses motion estimation only to guide reconstruction.
Within each group of frames, unchanged Gaussians keep their current versions in a shared dictionary, while modified Gaussians are appended as new versions.
To maximize version reuse, we introduce selective sequential construction, which refines the Gaussians selected by optical flow, photograph differences or rendering errors and leaves the others unchanged.
The dictionary is then pruned, fine-tuned with simulated quantization and encoded as images.
Adjusting the pruning ratio trades off storage and quality without rebuilding the dictionary.
On the fast dance sequences of DanceNet3D, VersionGS improves foreground PSNR by 2.7\,dB over the best baseline, and its PSNR decreases by less than 0.3\,dB on the highest-motion frames.
On N3DV and Technicolor, it requires 6.0$\times$ and 5.7$\times$ less storage than TAD-GS, the strongest baseline overall, while achieving even higher PSNR.
Finally, playback is fast and requires no motion model or neural network, only a 3DGS rasterizer.
\end{abstract}

%% file: sec/1_intro.tex
\begin{figure*}[t]
  \centering
  \includegraphics[width=0.963\textwidth]{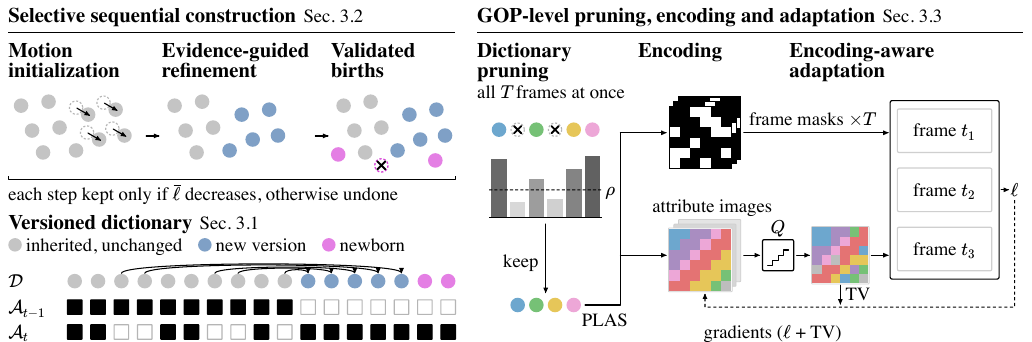}
  \caption{\textbf{Overview.} \emph{Left:} selective sequential construction in \cref{sec:construction} builds each frame from the previous one. In the dictionary $\mathcal D$ of \cref{sec:dictionary}, a changed Gaussian becomes a new version and an unchanged one adds only a reference. \emph{Right:} the GOP-level stages of \cref{sec:coding} are dictionary pruning, image-based encoding with a frame mask per version, and encoding-aware adaptation through the quantizer.}
  \label{fig:framework}
\end{figure*}

\section{Introduction}
\label{sec:intro}

Free-viewpoint video allows viewers to explore captured dynamic events from novel viewpoints. Reconstructing such scenes requires capturing fast motion, deforming clothing, and objects entering or leaving the view, while keeping the representation compact for streaming and lightweight to decode on everyday devices. While 3D Gaussian Splatting (3DGS)~\cite{kerbl3Dgaussians} enables real-time rendering of static scenes, and recent image-based formats bring compressed static Gaussian scenes to web and mobile viewers~\cite{morgenstern2024compact,playcanvas2025sog}, extending 3DGS to dynamic scenes requires an effective representation of how the Gaussians change over time.

Existing methods model temporal changes in two primary ways. One group models the change of each Gaussian as a smooth function of time, through deformation fields over a canonical set~\cite{Wu_2024_CVPR,yang2024deformable,huang2024scgs,ke2026streamstgs}, time-conditioned Gaussians~\cite{yang2023gs4d,Li_STG_2024_CVPR,sandu2026tadgs} or keyframe trajectories~\cite{lee2024ex4dgs}. Frame-to-frame methods instead update the Gaussians of the previous frame with per-frame transformations~\cite{sun20243dgstream,gao2024hicom,Hu_2025_CVPR}, quantized residuals~\cite{girish2024queen} or sparse control-point motion~\cite{zhang2025dfcgs}.

\looseness = -1 However, both groups lose quality as the magnitude of the motion increases. On fast dance sequences~\cite{wei2026dancenet3d}, these methods suffer a 1.1--4.0\,dB PSNR drop on the frames with the highest motion compared to those with the least (\cref{fig:teaser,fig:motion}). In temporally shared Gaussian representations, fast motion must be modeled either by persistent Gaussians that explain all timestamps or by transient Gaussians with short temporal lifespans, and neither works well. Optimizing a persistent Gaussian frame by frame under large inter-frame displacements degrades other frames~\cite{kim2026implicit}, whereas transient Gaussians with short lifespans are rarely sampled for densification, leaving dynamic regions under-reconstructed or blurry~\cite{sandu2026tadgs}. Meanwhile, frame-to-frame methods' position gradients are computed only within a limited local neighborhood and become inaccurate when objects move outside it, so Gaussians fade or are incorrectly recolored~\cite{trackersplat} and large inter-frame motion is underestimated~\cite{lee2026morgs}.

\looseness=-1 To avoid these problems, we propose VersionGS, which stores the Gaussian states of each frame rather than modeling their change. Within each group of frames (GOP), Gaussians that do not change are reused, a Gaussian that changes is stored again as a new version, and all versions form a dictionary shared by the frames of the GOP (\cref{fig:framework}). A changing Gaussian never has to explain all timestamps, and faster motion increases storage rather than degrading quality.

\looseness=-1 We build this dictionary offline, frame by frame, with selective sequential construction. Multi-view optical flow~\cite{kroeger2016dis} identifies moving Gaussians and translates them to follow large motion. These and additionally selected Gaussians are further refined to recover residual error, while newly appearing content is captured by spawning validated new Gaussians (\cref{fig:selectors}). All remaining Gaussians keep their exact prior versions, since refining them as well would slightly perturb static primitives and spawn redundant versions, which multiplies the storage without improving quality. For efficient transmission, we adapt static 3DGS compression~\cite{morgenstern2024compact,morgenstern2026kissgs} to each GOP dictionary: we prune versions by their contribution across frames, fine-tune them with simulated quantization and encode attributes into 2D images. Varying the ratio of kept versions adjusts the storage without rebuilding the dictionary. At playback, each GOP is decoded once, and each frame gathers its active versions via a binary mask for a standard 3DGS rasterizer, completely bypassing motion models or neural networks.

Our contributions are as follows:
\begin{itemize}\setlength{\itemsep}{0pt}\setlength{\parskip}{0pt}
\item \textbf{Per-frame states over motion models.} VersionGS encodes explicit per-frame Gaussian states in a shared dictionary, reusing static Gaussians exactly while appending modified ones as new versions, so fast motion trades off storage rather than visual quality.
\item \textbf{Compact by construction.} Selective sequential construction updates only Gaussians with clear evidence of change, cutting storage by up to 9.6$\times$ over full-frame refinement, while GOP-level pruning and image-based encoding enable flexible rate adjustment without rebuilding.
\item \textbf{Motion-robust quality at low bitrate.} VersionGS improves foreground PSNR on DanceNet3D by 2.7\,dB over the strongest baseline with under 0.3\,dB degradation on the fastest frames, reduces storage by 6.0$\times$ on N3DV~\cite{li2022n3dv} and 5.7$\times$ on Technicolor~\cite{sabater2017technicolor} over TAD-GS at higher PSNR, and renders using only a standard 3DGS rasterizer.
\end{itemize}

%% file: sec/2_related.tex
\section{Related Work}
\label{sec:related}

\input{fig/failures_chimee}
\paragraph{Temporal Gaussian representations.} Building on 3DGS~\cite{kerbl3Dgaussians}, \emph{deformation-based methods} transform canonical Gaussians with HexPlane features and a small decoder~\cite{Wu_2024_CVPR}, a time conditioned MLP~\cite{yang2024deformable} or sparse control points~\cite{huang2024scgs}. \emph{Space-time methods} make primitives functions of time, as in native 4D Gaussians conditioned on time~\cite{yang2023gs4d}, Spacetime Gaussians with temporal opacity and polynomial motion~\cite{Li_STG_2024_CVPR}, and TAD-GS~\cite{sandu2026tadgs} with temporal visibility and offset warping. \emph{Trajectory methods} store poses per frame under a rigidity prior~\cite{luiten2024dynamic} or interpolate them between keyframes~\cite{lee2024ex4dgs}. Temporal Gaussian Hierarchy~\cite{xu2024tgh} shares Gaussians across segments where content changes little. SPIN-4DGS~\cite{kim2026implicit} predicts attributes from spatiotemporal positions to handle large displacements.

\paragraph{Sequential reconstruction.} Streaming methods sequentially adapt Gaussians inherited from the previous frame. 3DGStream~\cite{sun20243dgstream} predicts translations and rotations with a hash-grid cache and adds Gaussians for new content. HiCoM~\cite{gao2024hicom}, ReCon-GS~\cite{recongs2025} and MoRGS~\cite{lee2026morgs} model the motion hierarchically or with flow supervision, and QUEEN~\cite{girish2024queen} learns residuals of all attributes with gated, selective updates. Population control from static reconstruction, by splitting, cloning and pruning~\cite{kerbl3Dgaussians} or by MCMC relocation under a budget~\cite{kheradmand2024mcmc}, covers content these updates cannot. Like QUEEN and PackUV~\cite{rai2026packuv}, which freezes static Gaussians identified by optical flow and fits attributes in UV atlases, we update only selected Gaussians, but we store each changed Gaussian as an unrestricted new version rather than as a residual.

\paragraph{Static Gaussian compression.} Static methods combine \emph{compaction}, which removes Gaussians~\cite{fan2024lightgaussian,fang2024minisplatting,kheradmand2024mcmc}, \emph{encoding}, which lowers the bits per Gaussian by vector quantization~\cite{navaneet2024compgs,lee2024compact,niedermayr2024compressed} or learned context models~\cite{lu2024scaffold,chen2024hac,chen2025hacpp,fcgs2025}, and \emph{adaptation} of the parameters to the codec~\cite{bagdasarian2025zip,morgenstern2026kissgs}. Learned coders reach high compression ratios, but their networks must run at decoding. Image-based layouts~\cite{morgenstern2024compact,morgenstern2026kissgs} instead sort the Gaussians into smooth attribute images for standard image codecs, so that decoding reduces to image decompression, and web viewers such as SuperSplat already use them~\cite{playcanvas2025sog}. Layered models add progressive streaming~\cite{shi2025lapisgs}. We build on the image-based layout.

\paragraph{Dynamic Gaussian compression.} Dynamic methods predict each frame from the previous one, code a temporal parameterization, or code frames jointly. Prediction-based methods, namely 4DGC~\cite{Hu_2025_CVPR} with its progressive extension~\cite{zheng20254dgcpro}, D-FCGS~\cite{zhang2025dfcgs} and a binary transformation cache~\cite{tang2025compressing}, reach low rates, but their quality depends on how well the motion model predicts the next frame, and their decoders run networks. Parameterization-based methods compress keyframe trajectories~\cite{lee2025rd4dgs}, or code canonical attributes as images and temporal features as video~\cite{ke2026streamstgs,wang2024v3}. Concurrently, Wang \etal~\cite{wang2026robust} concatenate per-frame Gaussian sets and code them with a static codec, with large gains only when the per-frame estimator tracks Gaussians across frames. Our reconstruction is instead built to preserve exact state reuse while allowing unrestricted new versions.

%% file: fig/failures_chimee.tex
\begin{figure}[t]
  \centering
  \scriptsize
  \setlength{\tabcolsep}{0.5pt}
  \renewcommand{\arraystretch}{0.4}
  \begin{tabular}{ccccc}
    \textbf{Ground truth} & \shortstack{\textbf{VersionGS}\\[1pt]$\rho=1$} & \shortstack{\textbf{VersionGS}\\[1pt]$\rho=0.35$} & \textbf{TAD-GS} & \textbf{4DGS} \\[2pt]
    \includegraphics[width=0.194\linewidth]{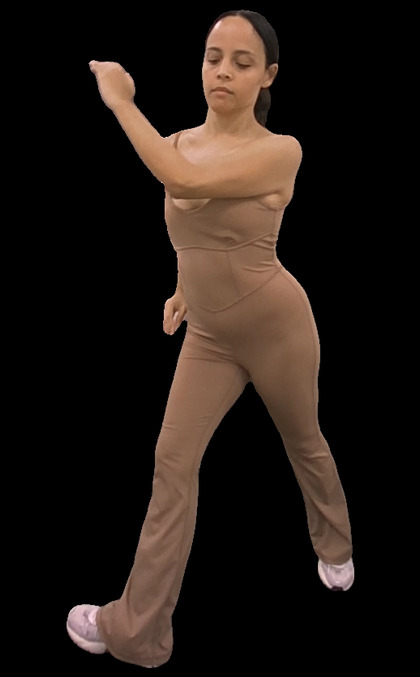} & \includegraphics[width=0.194\linewidth]{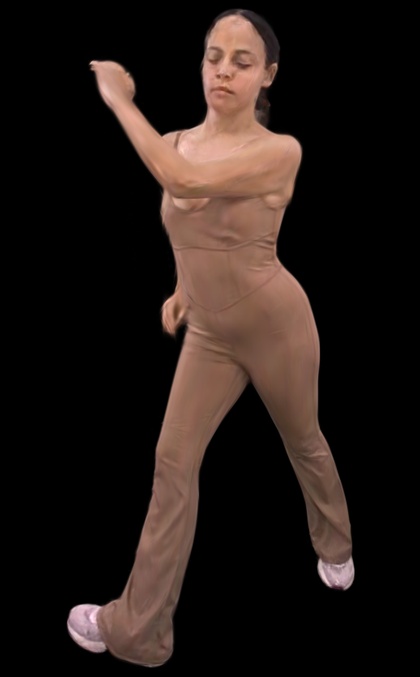} & \includegraphics[width=0.194\linewidth]{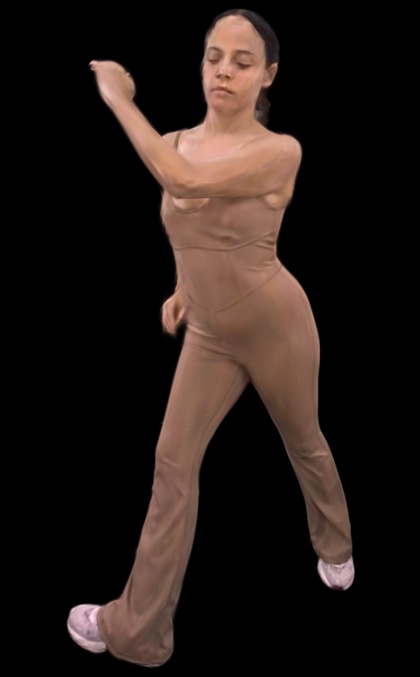} & \includegraphics[width=0.194\linewidth]{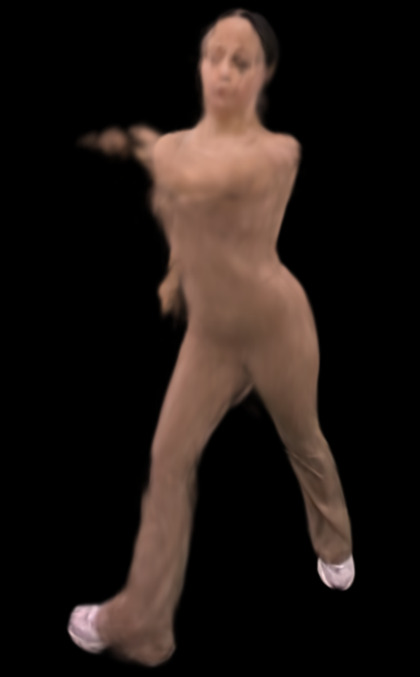} & \includegraphics[width=0.194\linewidth]{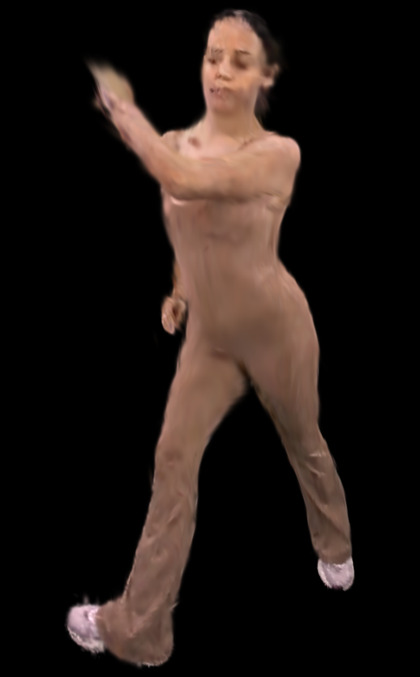} \\
    & 25.46\,dB & 25.52\,dB & 17.42\,dB & 22.24\,dB \\[4pt]
    \textbf{MoRGS} & \textbf{StreamSTGS} & \textbf{Ex4DGS} & \textbf{D-FCGS} & \textbf{QUEEN} \\[2pt]
    \includegraphics[width=0.194\linewidth]{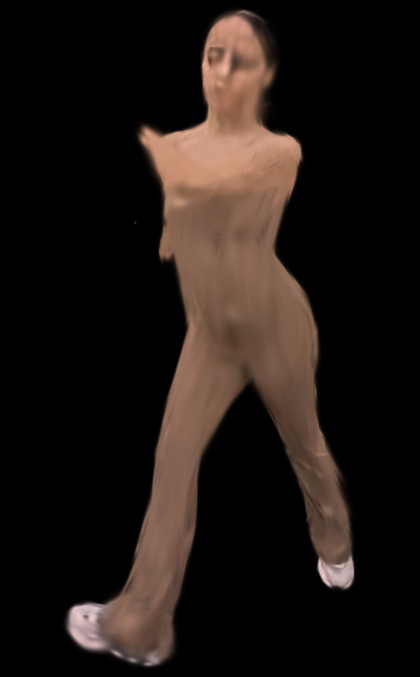} & \includegraphics[width=0.194\linewidth]{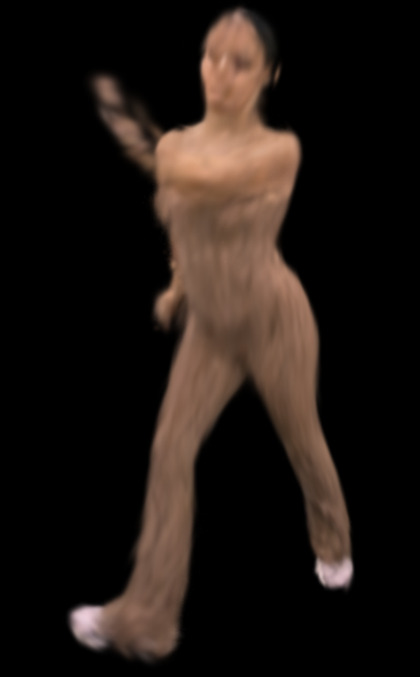} & \includegraphics[width=0.194\linewidth]{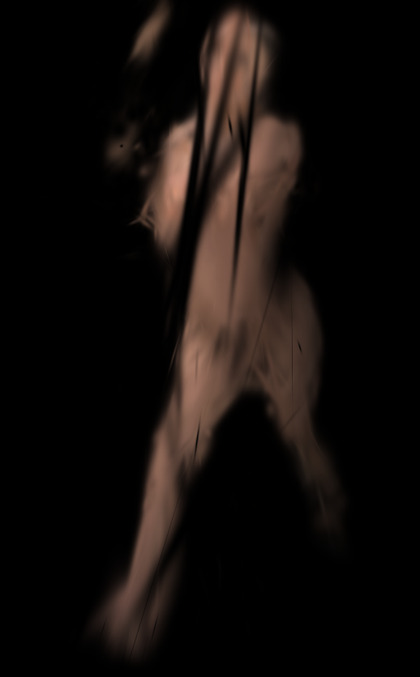} & \includegraphics[width=0.194\linewidth]{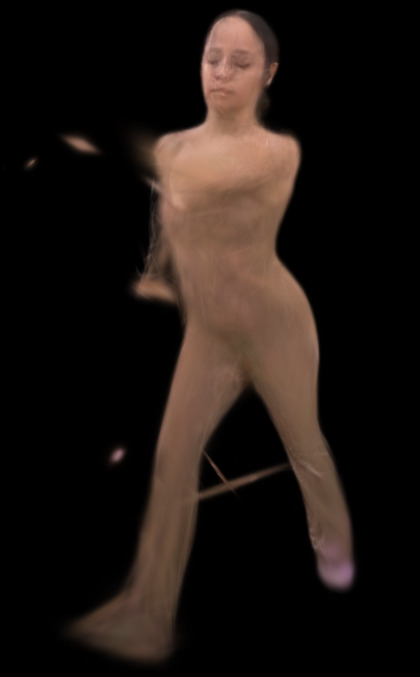} & \includegraphics[width=0.194\linewidth]{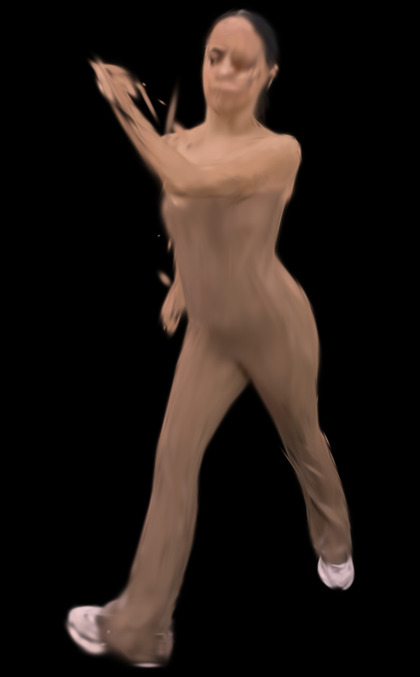} \\
    16.92\,dB & 17.88\,dB & 10.77\,dB & 15.30\,dB & 22.75\,dB \\
  \end{tabular}
  \caption{\textbf{Failure cases on DanceNet3D} in frame 150 of chimee. The baselines blur, smear or lose the moving performer. PSNR is computed inside the person mask.}
  \label{fig:failures}
\end{figure}

%% file: sec/3_method.tex
\section{Method}
\label{sec:method}
\input{fig/selectors}
Given a calibrated multi-view video, we split it into GOPs and represent each GOP by a dictionary of complete Gaussian versions and per-frame references, as \cref{fig:framework} shows. Selective sequential construction builds the dictionary frame by frame, and GOP-level pruning, image-based encoding and encoding-aware adaptation then compact and compress the dictionary for efficient streaming.

\subsection{Versioned Gaussian Dictionary}
\label{sec:dictionary}

A \emph{version} $g_j$ stores a complete set of standard 3DGS attributes. A GOP of $T$ frames is represented by a single dictionary $\mathcal{D}$ of $K$ versions, where each frame $t$ is defined by the subset of version indices $\mathcal{A}_t$ it activates:
\begin{equation}
  \mathcal D=\{g_j\}_{j=1}^{K},\qquad
  \mathcal A_t\subseteq\{1,\dots,K\},\qquad
  G_t=\{g_j\}_{j\in\mathcal A_t}.
  \label{eq:dictionary}
\end{equation}
A version is referenced at most once per frame, with the same attributes across every frame that uses it. At playback, the client decodes the GOP dictionary once and renders frame $t$ simply by gathering the active set $G_t$ for a standard 3DGS rasterizer~\cite{kerbl3Dgaussians}. Because this requires no neural network evaluation, existing image-based web viewers~\cite{playcanvas2025sog} only need a lightweight mask decoder and an index-gather step to support dynamic playback.

During construction, any unchanged Gaussian from frame $t-1$ to $t$ simply retains its dictionary index in $\mathcal{A}_t$.  If a Gaussian is modified, its updated state is appended to $\mathcal{D}$ as a new version referenced by $\mathcal{A}_t$, while earlier frames continue to reference the prior version. Newly introduced Gaussians similarly append new versions to $\mathcal{D}$. Construction never changes previously committed versions or frame masks. Once construction is complete, the GOP-level stages of \cref{sec:coding} prune less important versions and jointly fine-tune the retained dictionary, separating reconstruction from coding.

\subsection{Selective Sequential Construction}
\label{sec:construction}

We initialize the dictionary at the first frame of the GOP using MCMC-based 3DGS~\cite{kheradmand2024mcmc}. Each subsequent frame $t$ inherits the active Gaussians $G_{t-1}$ and goes through four selective update stages: motion initialization, evidence-guided refinement, residual recovery and validated births. Any Gaussian not modified by these stages continues to reference its version from frame $t-1$. Every stage is guarded by a strict validation check that an update is committed only if it reduces the multi-view reconstruction loss. \Cref{fig:selectors} illustrates which Gaussians each stage updates.

\paragraph{Reconstruction objective.}
All optimization stages minimize a combination of $L_1$ and SSIM losses
\begin{equation}
\begin{aligned}
  \hat I_v^b &= C_v + (1-A_v)\,b, \\
  \ell(G,v,b) &= 0.8\,\lVert \hat I_v^b-I_v^b\rVert_1 \\
  &\quad + 0.2\,\bigl[1-\mathrm{SSIM}(\hat I_v^b,I_v^b)\bigr],
\end{aligned}
\label{eq:loss}
\end{equation}
where $\lVert\cdot\rVert_1$ is the mean absolute pixel error, $C_v$ and $A_v$ are the premultiplied color and alpha rendered from Gaussian set $G$ in training view $v$, $b$ is the background color, and $I_v^b$ is the target training image or, for foreground-only sequences, the masked foreground composited over $b$. During fitting, $b$ is sampled uniformly from $[0,1]^3$ to penalize unexplained transparency~\cite{mueller2022instant,tancik2023nerfstudio,wei2026dancenet3d}. To make acceptance decisions deterministic, we evaluate candidates against both black and white backgrounds
\begin{equation}
  \bar\ell(G,\mathcal V)=\frac{1}{2\lvert\mathcal V\rvert}\sum_{v\in\mathcal V}\,\sum_{b\in\{\mathbf 0,\mathbf 1\}}\ell(G,v,b),
  \label{eq:decision}
\end{equation}
where $\bar\ell$ denotes the loss over all training views of the current frame. For foreground-only sequences, the $L_1$ error is normalized by the foreground area and SSIM~\cite{wang2004ssim} is computed in the foreground bounding box expanded by 16 pixels (8 pixels during evaluation).

\input{fig/rd_curves}\paragraph{Motion initialization.}
\looseness=-1 To handle large inter-frame displacements, we compute dense 2D optical flow~\cite{kroeger2016dis} between frames $t-1$ and $t$ in all training views. Sampled at projected Gaussian centers, it gives a 3D displacement per Gaussian by weighted least squares~\cite{vedula1999sceneflow}. Gaussians with a clear displacement are grouped spatially, and we optimize a single rigid 3D translation per group against the training images of frame $t$. A group is subdivided if splitting further reduces $\bar{\ell}$, allowing the motion granularity to adapt to the scene.

\paragraph{Evidence-guided refinement.}
\leavevmode Optimizing every inherited Gaussian would slightly perturb static primitives, turning reusable dictionary entries into redundant new versions. We therefore restrict gradient refinement strictly to Gaussians showing clear evidence of change via two selectors. The \emph{motion selector} picks the Gaussians that motion initialization moved. The \emph{image-change selector} picks the Gaussians whose image contribution falls largely on changed pixels, measured against a per-camera reference image that, like the frame memory of conditional replenishment~\cite{mounts1969conditional}, lets slow changes add up until they are detected. We then optimize the selected Gaussians' centers, scales, rotations, opacities and DC colors, with higher-order SH fixed, committing the updated versions only if it lowers $\bar\ell$ compared with simply inheriting frame $t-1$.

\paragraph{Residual recovery.}
To capture appearance changes missed by the initial selectors, two short refinement passes select a limited number of Gaussians whose contribution to the image falls largely on high-error pixels. A short first pass tests which errors a refit can remove. Pixels with high error before and after the first pass form the regions used by births. A longer second pass then refits the Gaussians in the remaining high error regions. The version with the lowest $\bar\ell$ is kept.

\paragraph{Validated births.}
A birth adds Gaussians for content that frame $t-1$ lacks, such as a person entering the view. In regions where image changes and rendering errors persist, we propose candidate new Gaussians via multi-view plane-sweep depth~\cite{collins1996space} and optimize them on a subset of training views with existing Gaussians fixed. A candidate group is accepted if it improves the training views held back from fitting, passes per-view error checks, and reduces the loss enough to justify its estimated storage~\cite{sullivan1998rdo}. Accepted primitives are appended to $\mathcal{D}$ as new versions.

\subsection{GOP-Level Pruning, Encoding and Adaptation}
\label{sec:coding}
\input{sec/tab_main}

\paragraph{Dictionary pruning.}
Once all $T$ frames of a GOP are constructed, we compact the dictionary $\mathcal{D}$ with the GaussianPOP \emph{leave-one-out} score~\cite{lee2026gaussianpop}, following KISS-GS~\cite{morgenstern2026kissgs}. Let $C_{t,v}$ be the render of frame $t$ in training view $v$ and $C_{t,v}^{\setminus j}$ the same render without version $j$. Each version is scored by the squared change of the renders when it is removed, summed over all frames and views in which it is active, 
\begin{equation}
  s_{j,t}=\sum_{v}\bigl\lVert C_{t,v}-C_{t,v}^{\setminus j}\bigr\rVert_2^2,\qquad
  s_j=\sum_{t:\,j\in\mathcal A_t}s_{j,t},
  \label{eq:prune_score}
\end{equation}
GaussianPOP computes $s_{j,t}$ for all active versions simultaneously in a single compositing pass per frame. Given a target keep fraction $\rho \in (0, 1]$, we prune the $(1 - \rho)K$ versions with the lowest total score $s_j$ from $\mathcal{D}$ and all frame masks $\mathcal{A}_t$.
Versions that follow fast motion are active for only a few frames, so their summed scores are smaller than those of persistent background Gaussians, and pruning by $s_j$ alone removes moving content first. We therefore weight the part of each score that falls on moving pixels. With $\Delta_{t,v}=C_{t,v}-C_{t,v}^{\setminus j}$, the per-view term of \cref{eq:prune_score} becomes
\begin{equation}
  \bigl\lVert (1-M_{t,v})\odot\Delta_{t,v}\bigr\rVert_2^2+\beta\,\bigl\lVert M_{t,v}\odot\Delta_{t,v}\bigr\rVert_2^2,
  \label{eq:prune_motion}
\end{equation}
where the mask $M_{t,v}$ marks the pixels of training view $v$ whose color changes between frames $t-1$ and $t$, and $\beta=32$. Varying $\rho$ adjusts the bitrate by re-running only pruning, adaptation, and encoding, not the construction.

\paragraph{Encoding.}
Following Self-Organizing Gaussian Grids~\cite{morgenstern2024compact} and KISS-GS~\cite{morgenstern2026kissgs}, we map the versions onto 2D grids using Parallel Linear Assignment Sorting (PLAS)~\cite{morgenstern2024compact} based on their quantized centers. Positions, scales, rotations, opacity and DC color are quantized into integer attribute images, while higher-order SH coefficients are vector-quantized into a codebook with an image of labels. All images, along with an occupancy mask for valid cells, are compressed losslessly with WebP~\cite{google2010webp}. We treat frame membership as a $T$-bit mask per version in the same order, coded with an arithmetic coder~\cite{witten1987arithmetic}. Our reported storage includes all images, codebooks, masks, coding tables and metadata.

\paragraph{Encoding-aware adaptation.}
\looseness=-1 Construction fits versions one frame at a time, yet a shared version serves several frames and quantization introduces error to attributes. We therefore refine the dictionary jointly over all $T$ frames of the GOP. Keeping frame masks $\mathcal{A}_t$, the 2D grid layout, and SH codebook assignments fixed, we improve quality, adapt the model to quantization error, and further increase the compressibility of the attributes by solving:
\begin{equation}
\begin{aligned}
  \min_{\mathcal D}\quad &
  \frac{1}{T}\sum_{t=1}^{T}\frac{1}{\lvert\mathcal V\rvert}\sum_{v\in\mathcal V}
  \mathbb E_b\bigl[\ell(Q(\mathcal D)[\mathcal A_t],v,b)\bigr] \\
  &+\lambda_{\mathrm{TV}}\,\mathcal R_{\mathrm{TV}}\bigl(Q(\mathcal D)\bigr),
\end{aligned}
\label{eq:adaptation}
\end{equation}
where $Q$ simulates decoder quantization with the straight-through estimator~\cite{bengio2013estimating}, and $\mathcal R_{\mathrm{TV}}$ is the total-variation (TV) regularizer of KISS-GS~\cite{morgenstern2026kissgs}, the mean absolute difference between adjacent occupied grid cells. KISS-GS's Parallel Representative Assignment Smoothing (PRAS) also selects equivalent rotation--scale parameterizations that smooth the grids without changing the Gaussians' shapes. Finally, we average the parameters of several late checkpoints~\cite{izmailov2018averaging} and encode the result.

%% file: fig/selectors.tex
\definecolor{selMotion}{rgb}{0.00,0.45,0.85}\definecolor{selPhoto}{rgb}{1.00,0.60,0.00}%
\definecolor{selRecovery}{rgb}{0.10,0.75,0.30}\definecolor{selBirth}{rgb}{1.00,0.10,0.90}%
\newcommand{\selkey}[1]{\textcolor{#1}{\rule{0.6em}{0.6em}}}%
\begin{figure}[t]
  \centering
  \includegraphics[width=\columnwidth]{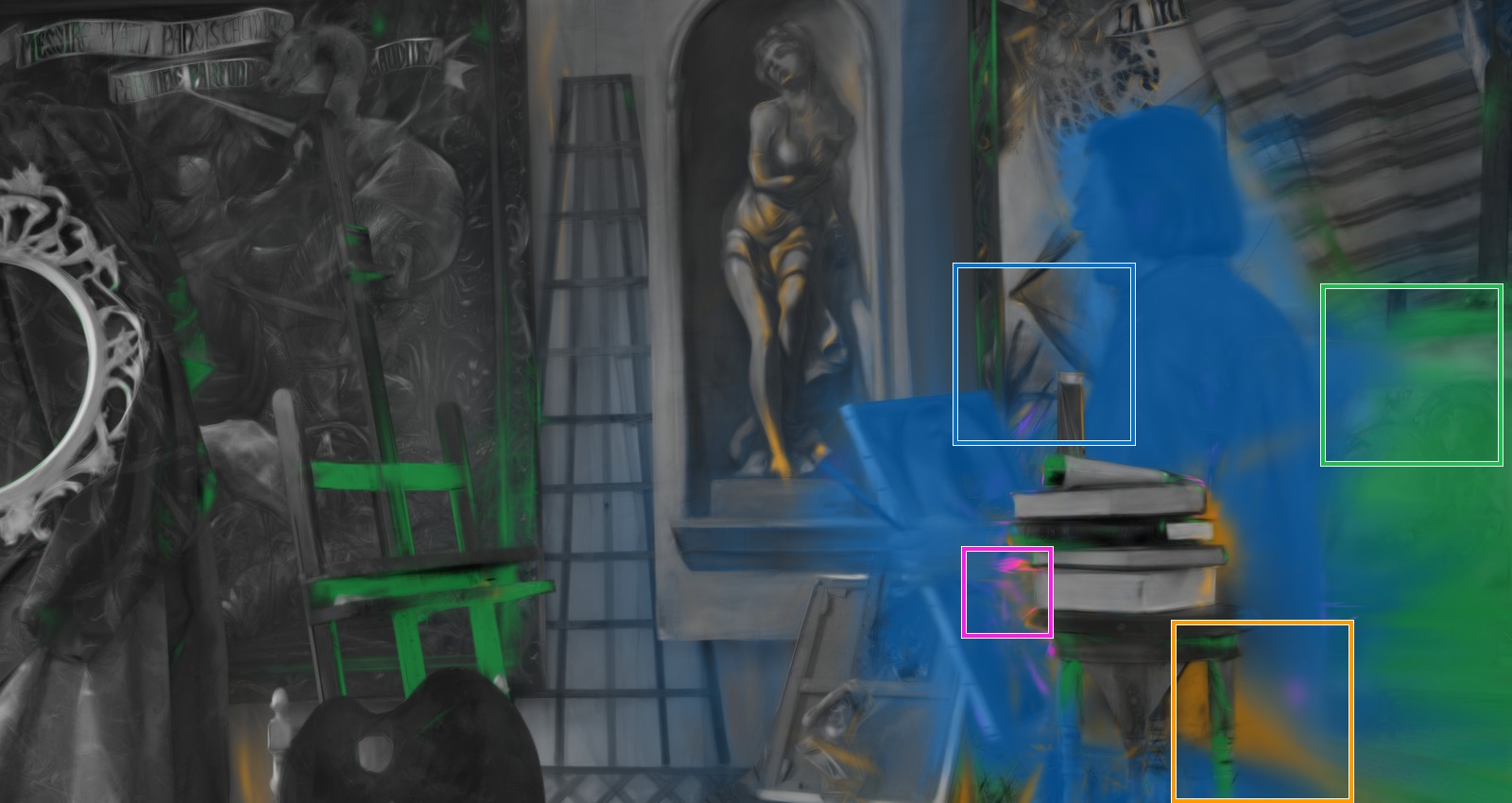}\\[0.01\columnwidth]
  \includegraphics[width=0.244\columnwidth]{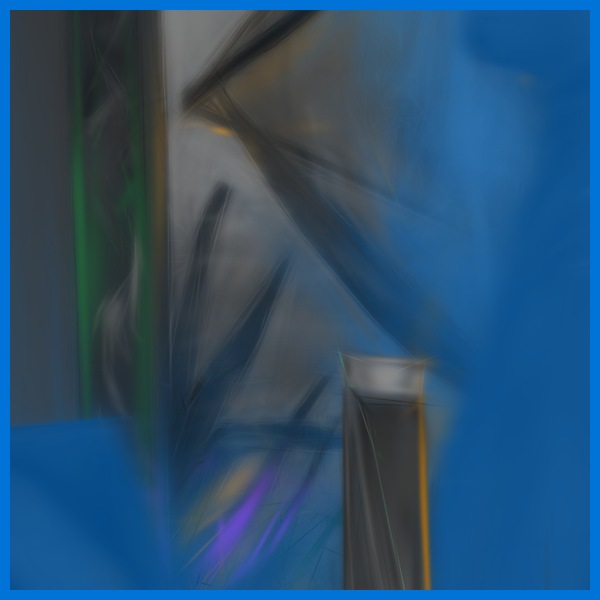}\hfill
  \includegraphics[width=0.244\columnwidth]{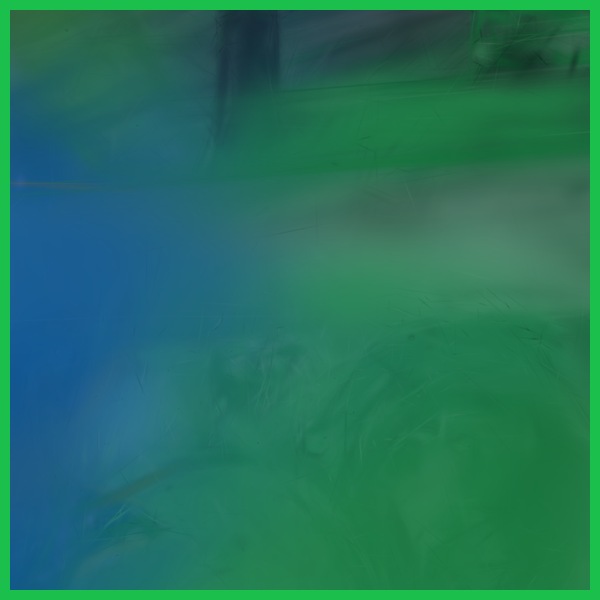}\hfill
  \includegraphics[width=0.244\columnwidth]{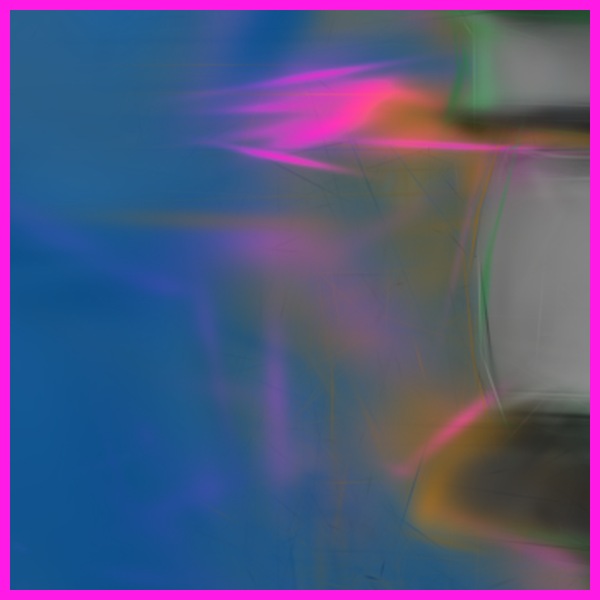}\hfill
  \includegraphics[width=0.244\columnwidth]{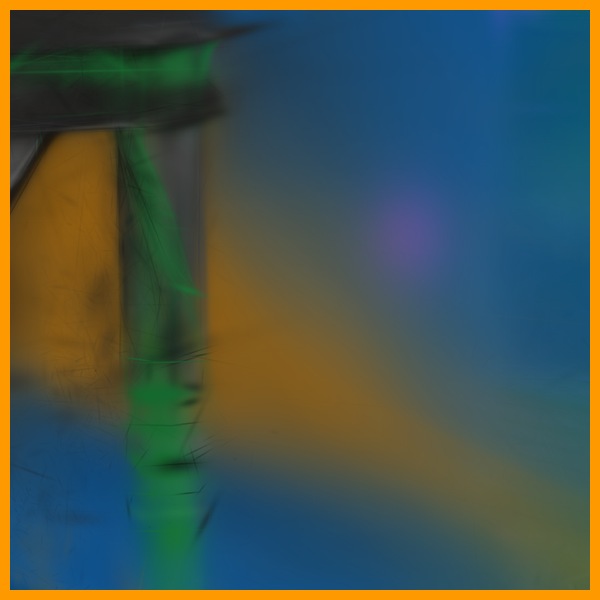}
  \caption{\textbf{What one construction step changes} in a held-out view of painter, frame $40\to41$. The person moves from right to left, and the selectors pick the Gaussians on the moving person and in the previously occluded areas. Colored Gaussians became new versions through the \selkey{selMotion}~motion selector, \selkey{selPhoto}~photograph-change selector, \selkey{selRecovery}~residual recovery or \selkey{selBirth}~validated births. 22,157 of 304,021 Gaussians change in this step, about 7\%.}
  \label{fig:selectors}
\end{figure}

%% file: fig/rd_curves.tex
\definecolor{rdC1}{HTML}{8C564B}\definecolor{rdC2}{HTML}{9467BD}\definecolor{rdC3}{HTML}{7F7F7F}\definecolor{rdC4}{HTML}{17BECF}%
\definecolor{rdC5}{HTML}{BCBD22}\definecolor{rdC6}{HTML}{1F77B4}\definecolor{rdC7}{HTML}{2CA02C}\definecolor{rdC8}{HTML}{FF7F0E}\definecolor{rdOurs}{HTML}{D62728}%
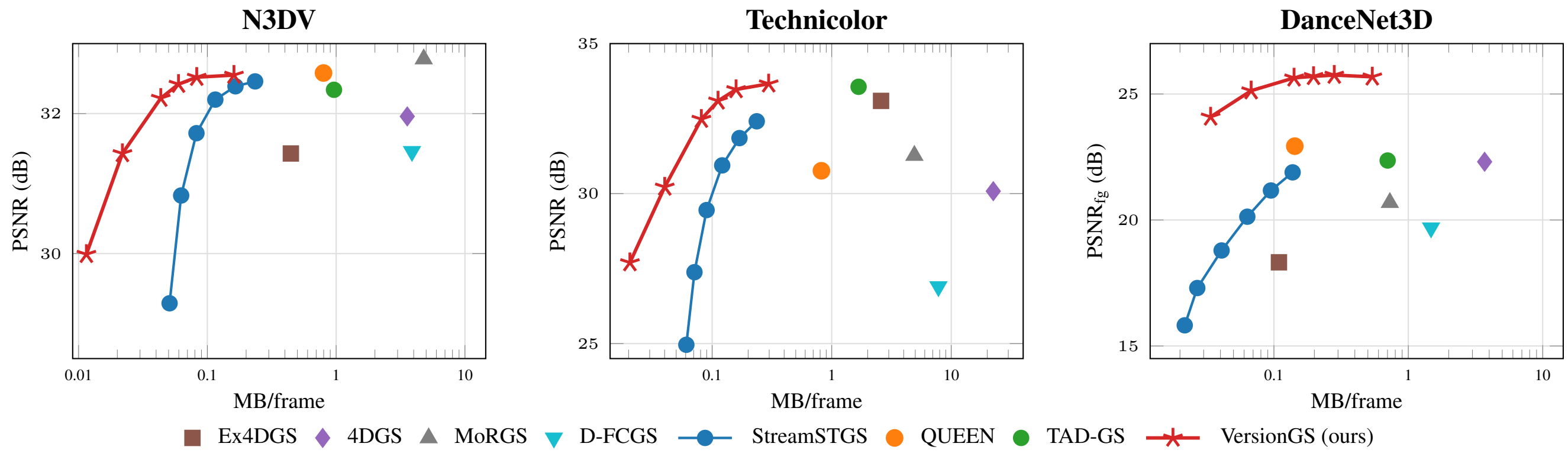
\begin{figure*}[t]
  \centering
  \begin{tikzpicture}\begin{axis}[width=0.355\textwidth, height=5.1cm, xmode=log, log basis x=10, grid=major, grid style={gray!25}, xlabel={MB/frame}, xlabel style={font=\scriptsize, yshift=3pt}, ylabel style={font=\scriptsize, yshift=-4pt}, tick label style={font=\tiny}, title style={font=\small\bfseries, yshift=-5pt}, clip mode=individual, xmin=0.009, xmax=14.504, xtick={0.01,0.1,1,10}, xticklabels={0.01,0.1,1,10}, minor xtick={0.009,0.02,0.03,0.04,0.05,0.06,0.07,0.08,0.09,0.2,0.3,0.4,0.5,0.6,0.7,0.8,0.9,2,3,4,5,6,7,8,9}, title={N3DV}, ylabel={PSNR (dB)}, ymin=28.5, ymax=33.0, legend to name=rdlegend, legend columns=8, legend style={font=\scriptsize, draw=none, column sep=4pt}]
    \addplot[line width=0.8pt, mark size=2.2pt, color=rdC1, mark=square*, only marks] coordinates {(0.445,31.43)};
    \addlegendentry{Ex4DGS}
    \addplot[line width=0.8pt, mark size=2.2pt, color=rdC2, mark=diamond*, mark size=2.6pt, only marks] coordinates {(3.567,31.96)};
    \addlegendentry{4DGS}
    \addplot[line width=0.8pt, mark size=2.2pt, color=rdC3, mark=triangle*, mark size=2.6pt, only marks] coordinates {(4.787,32.78)};
    \addlegendentry{MoRGS}
    \addplot[line width=0.8pt, mark size=2.2pt, color=rdC4, mark=triangle*, mark options={rotate=180}, mark size=2.6pt, only marks] coordinates {(3.881,31.46)};
    \addlegendentry{D-FCGS}
    \addplot[line width=0.8pt, mark size=2.2pt, color=rdC6, mark=*] coordinates {(0.0509,29.29) (0.0626,30.83) (0.0823,31.72) (0.1153,32.20) (0.1653,32.39) (0.2349,32.46)};
    \addlegendentry{StreamSTGS}
    \addplot[line width=0.8pt, mark size=2.2pt, color=rdC8, mark=oplus*, mark size=2.4pt, only marks] coordinates {(0.798,32.58)};
    \addlegendentry{QUEEN}
    \addplot[line width=0.8pt, mark size=2.2pt, color=rdC7, mark=otimes*, only marks] coordinates {(0.962,32.34)};
    \addlegendentry{TAD-GS}
    \addplot[line width=0.8pt, mark size=2.2pt, color=rdOurs, mark=star, mark size=3.2pt, line width=1.1pt, mark options={line width=0.9pt}] coordinates {(0.0115,29.99) (0.0219,31.43) (0.0435,32.22) (0.0597,32.42) (0.0827,32.52) (0.1609,32.55)};
    \addlegendentry{VersionGS (ours)}
  \end{axis}\end{tikzpicture}%
  \hfill
  \begin{tikzpicture}\begin{axis}[width=0.355\textwidth, height=5.1cm, xmode=log, log basis x=10, grid=major, grid style={gray!25}, xlabel={MB/frame}, xlabel style={font=\scriptsize, yshift=3pt}, ylabel style={font=\scriptsize, yshift=-4pt}, tick label style={font=\tiny}, title style={font=\small\bfseries, yshift=-5pt}, clip mode=individual, xmin=0.014, xmax=39.961, xtick={0.1,1,10}, xticklabels={0.1,1,10}, minor xtick={0.02,0.03,0.04,0.05,0.06,0.07,0.08,0.09,0.2,0.3,0.4,0.5,0.6,0.7,0.8,0.9,2,3,4,5,6,7,8,9,20,30}, title={Technicolor}, ylabel={PSNR (dB)}, ymin=24.5, ymax=35.0]
    \addplot[line width=0.8pt, mark size=2.2pt, color=rdC1, mark=square*, only marks] coordinates {(2.586,33.09)};
    \addplot[line width=0.8pt, mark size=2.2pt, color=rdC2, mark=diamond*, mark size=2.6pt, only marks] coordinates {(22.472,30.08)};
    \addplot[line width=0.8pt, mark size=2.2pt, color=rdC3, mark=triangle*, mark size=2.6pt, only marks] coordinates {(4.916,31.27)};
    \addplot[line width=0.8pt, mark size=2.2pt, color=rdC4, mark=triangle*, mark options={rotate=180}, mark size=2.6pt, only marks] coordinates {(7.812,26.9)};
    \addplot[line width=0.8pt, mark size=2.2pt, color=rdC6, mark=*] coordinates {(0.0608,24.96) (0.0711,27.38) (0.0898,29.45) (0.1213,30.94) (0.1692,31.85) (0.236,32.41)};
    \addplot[line width=0.8pt, mark size=2.2pt, color=rdC8, mark=oplus*, mark size=2.4pt, only marks] coordinates {(0.82,30.76)};
    \addplot[line width=0.8pt, mark size=2.2pt, color=rdC7, mark=otimes*, only marks] coordinates {(1.678,33.56)};
    \addplot[line width=0.8pt, mark size=2.2pt, color=rdOurs, mark=star, mark size=3.2pt, line width=1.1pt, mark options={line width=0.9pt}] coordinates {(0.0205,27.7) (0.0401,30.22) (0.0812,32.48) (0.112,33.09) (0.1585,33.47) (0.2968,33.66)};
  \end{axis}\end{tikzpicture}%
  \hfill
  \begin{tikzpicture}\begin{axis}[width=0.355\textwidth, height=5.1cm, xmode=log, log basis x=10, grid=major, grid style={gray!25}, xlabel={MB/frame}, xlabel style={font=\scriptsize, yshift=3pt}, ylabel style={font=\scriptsize, yshift=-4pt}, tick label style={font=\tiny}, title style={font=\small\bfseries, yshift=-5pt}, clip mode=individual, xmin=0.012, xmax=14.219, xtick={0.1,1,10}, xticklabels={0.1,1,10}, minor xtick={0.02,0.03,0.04,0.05,0.06,0.07,0.08,0.09,0.2,0.3,0.4,0.5,0.6,0.7,0.8,0.9,2,3,4,5,6,7,8,9}, title={DanceNet3D}, ylabel={PSNR$_\text{fg}$ (dB)}, ymin=14.5, ymax=27.0]
    \addplot[line width=0.8pt, mark size=2.2pt, color=rdC1, mark=square*, only marks] coordinates {(0.109,18.32)};
    \addplot[line width=0.8pt, mark size=2.2pt, color=rdC2, mark=diamond*, mark size=2.6pt, only marks] coordinates {(3.698,22.31)};
    \addplot[line width=0.8pt, mark size=2.2pt, color=rdC3, mark=triangle*, mark size=2.6pt, only marks] coordinates {(0.729,20.69)};
    \addplot[line width=0.8pt, mark size=2.2pt, color=rdC4, mark=triangle*, mark options={rotate=180}, mark size=2.6pt, only marks] coordinates {(1.477,19.69)};
    \addplot[line width=0.8pt, mark size=2.2pt, color=rdC6, mark=*] coordinates {(0.0217,15.82) (0.0269,17.3) (0.0408,18.79) (0.0634,20.13) (0.095,21.17) (0.1378,21.89)};
    \addplot[line width=0.8pt, mark size=2.2pt, color=rdC8, mark=oplus*, mark size=2.4pt, only marks] coordinates {(0.143,22.93)};
    \addplot[line width=0.8pt, mark size=2.2pt, color=rdC7, mark=otimes*, only marks] coordinates {(0.702,22.36)};
    \addplot[line width=0.8pt, mark size=2.2pt, color=rdOurs, mark=star, mark size=3.2pt, line width=1.1pt, mark options={line width=0.9pt}] coordinates {(0.0338,24.09) (0.0677,25.13) (0.1412,25.64) (0.1972,25.7) (0.2813,25.75) (0.5395,25.68)};
  \end{axis}\end{tikzpicture}%
  \\[-2pt]\pgfplotslegendfromname{rdlegend}
  \caption{\textbf{RD trade-off}, averaged over the scenes of each dataset. Our curve varies the pruning keep fraction $\rho\in\{0.06,0.12,0.25,0.35,0.50,1\}$ from left to right, and StreamSTGS varies the quality setting of its feature video. Other baselines cannot change rate without retraining. VersionGS shows consistent top performance across three tested datasets.}
  \label{fig:rd}
\end{figure*}

%% file: sec/tab_main.tex
\begin{table*}[t]
  \centering
  \setlength{\tabcolsep}{2.6pt}
  \resizebox{\textwidth}{!}{%
  \begin{tabular}{l ccccc ccccc ccccc}
    \toprule
    & \multicolumn{5}{c}{N3DV (6 scenes)} & \multicolumn{5}{c}{Technicolor (4 scenes)} & \multicolumn{5}{c}{DanceNet3D (4 scenes)} \\
    \cmidrule(lr){2-6}\cmidrule(lr){7-11}\cmidrule(lr){12-16}
    Method & PSNR$\uparrow$ & SSIM$\uparrow$ & LPIPS$\downarrow$ & MB/fr$\downarrow$ & FPS$\uparrow$
           & PSNR$\uparrow$ & SSIM$\uparrow$ & LPIPS$\downarrow$ & MB/fr$\downarrow$ & FPS$\uparrow$
           & PSNR$_\text{fg}$$\uparrow$ & SSIM$_\text{fg}$$\uparrow$ & LPIPS$_\text{fg}$$\downarrow$ & MB/fr$\downarrow$ & FPS$\uparrow$ \\
    \midrule
        Ex4DGS~\cite{lee2024ex4dgs}        & 31.43 & 0.943 & 0.095 & 0.445 & 83 & \rankC{33.09} & 0.940 & 0.055 & 2.586 & 67 & 18.32 & 0.778 & 0.376 & \rankB{0.109} & \rankB{528} \\
        4DGS~\cite{yang2023gs4d}            & 31.96 & 0.948 & \rankC{0.085} & 3.567 & \rankC{120} & 30.08$^\S$ & 0.869$^\S$ & 0.142$^\S$ & 22.472$^\S$ & 42$^\S$ & 22.31 & 0.842 & 0.282 & 3.698 & 204 \\
        MoRGS~\cite{lee2026morgs}     & \rankA{32.78} & \rankA{0.954} & \rankA{0.076} & 4.787 & 1 & 31.27 & 0.928 & 0.072 & 4.916 & 1 & 20.69 & 0.838 & 0.272 & 0.729 & 4 \\
        D-FCGS~\cite{zhang2025dfcgs}       & 31.46 & 0.945 & 0.099 & 3.881 & 5 & 26.90 & 0.883 & 0.103 & 7.812 & 2 & 19.69 & 0.805 & 0.306 & 1.477 & 18 \\
        StreamSTGS~\cite{ke2026streamstgs} & 32.39 & 0.946 & 0.093 & \rankC{0.165} & 26 & 31.85 & 0.923 & 0.075 & \rankB{0.169} & 26 & 21.17 & 0.831 & 0.303 & \rankA{0.095} & 46 \\
        QUEEN~\cite{girish2024queen}   & \rankB{32.58} & \rankC{0.952} & \rankB{0.084} & 0.798 & 1 & 30.76 & 0.920 & 0.079 & 0.820 & 1 & \rankC{22.93} & \rankC{0.855} & \rankC{0.240} & \rankC{0.143} & 5 \\
        TAD-GS~\cite{sandu2026tadgs}       & 32.34 & \rankB{0.953} & \rankA{0.076} & 0.962 & 98 & \rankB{33.56} & \rankB{0.948} & \rankB{0.045} & 1.678 & \rankC{75} & 22.36 & 0.851 & 0.271 & 0.702 & 121 \\
    \midrule
        Ours, $\rho{=}1$                 & \rankC{32.55} & \rankB{0.953} & \rankA{0.076} & \rankB{0.161} & \rankB{207} & \rankA{33.66} & \rankA{0.949} & \rankA{0.038} & \rankC{0.297} & \rankB{239} & \rankB{25.68} & \rankA{0.898} & \rankA{0.134} & 0.540 & \rankC{265} \\
        Ours, $\rho{=}0.35$                & 32.42 & 0.950 & 0.086 & \rankA{0.060} & \rankA{304} & \rankC{33.09} & \rankC{0.941} & \rankC{0.050} & \rankA{0.112} & \rankA{432} & \rankA{25.70} & \rankB{0.896} & \rankB{0.150} & 0.197 & \rankA{596} \\
    \bottomrule
  \end{tabular}}
  \caption{\textbf{Quantitative comparison} on the held-out cameras, averaged per dataset. On DanceNet3D, PSNR uses the person mask and SSIM and LPIPS its bounding box. FPS is end-to-end playback on one H100, decoding every frame from the stored files, and rendering one held-out view (see supplement). \protect\colorbox{red!25}{Best}, \protect\colorbox{orange!25}{second} and \protect\colorbox{yellow!35}{third} per column. $^\S$10M Gaussian cap. Our other $\rho$ in \cref{fig:rd}.}
  \label{tab:main}
\end{table*}

%% file: sec/4_experiments.tex
\section{Experiments}
\label{sec:experiments}

\subsection{Experimental Setup}
\label{sec:setup}

\paragraph{Datasets.}
We evaluate the first 300 frames per sequence, 290 for \emph{birthday}, on three multi-view datasets. N3DV~\cite{li2022n3dv}, a widely used dataset for dynamic Gaussian evaluation, contains six indoor scenes captured by 18--21 cameras at 30\,fps. We make sure all methods use identical frames, undistorted with the first-frame COLMAP calibration, with the central camera held out. Technicolor~\cite{sabater2017technicolor} contains light-field videos from a $4\times4$ camera rig, featuring complex texture and changing visibility. Following standard practice, we use \emph{birthday}, \emph{painter}, \emph{theater}, and \emph{train} at half resolution and hold out camera 10. DanceNet3D~\cite{wei2026dancenet3d} captures fast full-body motion and rapidly deforming clothing with 28--29 cameras. We evaluate \emph{chimee}, \emph{russia costume}, \emph{big kicks}, and \emph{biz markie}, reconstructing only the performer using person masks and holding out cameras 0879, 1040, and 1753.

\paragraph{Baselines.}
We compare VersionGS against seven dynamic Gaussian methods: keyframe-based Ex4DGS~\cite{lee2024ex4dgs}; space--time 4DGS~\cite{yang2023gs4d} and TAD-GS~\cite{sandu2026tadgs}; streaming MoRGS~\cite{lee2026morgs} and D-FCGS~\cite{zhang2025dfcgs}; and compressed streaming methods StreamSTGS~\cite{ke2026streamstgs} and QUEEN~\cite{girish2024queen}. TAD-GS serves as our primary visual quality benchmark, with the strongest baseline results on Technicolor and the best SSIM and LPIPS on N3DV. StreamSTGS, the most compact baseline, serves as our rate reference and supports rate adjustment without retraining. We use QP 20 in \cref{tab:main}. We use official implementations and published configurations, adopting N3DV training settings for datasets not covered by the original papers. On N3DV, all methods share preprocessed frames, camera calibration, and held-out evaluation photographs. On DanceNet3D, baselines are trained on the person composited on black, except StreamSTGS on \emph{biz markie}, which uses white. D-FCGS codes 3DGStream~\cite{sun20243dgstream} started from our first frame. On Technicolor, we cap 4DGS at 10M Gaussians to prevent out-of-memory errors.

\paragraph{Metrics.}
A unified evaluation script scores all methods on held-out views using PSNR, SSIM~\cite{wang2004ssim}, and LPIPS~\cite{zhang2018lpips} with AlexNet, computed per image and averaged across views and frames. On DanceNet3D, we instead compute PSNR within the person mask and SSIM and LPIPS within its bounding box expanded by 8 pixels. Temporal consistency is measured by tLP~\cite{chu2020tecogan}. It computes the absolute difference between the LPIPS distance of consecutive renders and that of the corresponding photographs. Storage is the total data required for decoding divided by the number of frames, reported in MB/frame. For VersionGS, this includes all attribute images, codebooks, frame masks, and metadata, with the first frame of every GOP included. Baselines are measured in their released model or bitstream formats.

\paragraph{VersionGS Settings.}
\leavevmode We use GOPs of $T=60$ frames, five per sequence, each coded independently. The first frame of each GOP is reconstructed with MCMC-based 3DGS~\cite{kheradmand2024mcmc}, using 150k Gaussians and 20k iterations on N3DV, 300k Gaussians and 90k iterations on Technicolor, and 50k Gaussians and 7k iterations on DanceNet3D. \Cref{tab:main} reports $\rho=1$, without pruning, and the balanced rate-distortion setting $\rho=0.35$. \Cref{fig:rd} adds $\rho\in\{0.50,0.25,0.12,0.06\}$.

\subsection{Comparison with Existing Methods}
\label{sec:comparison}

\Cref{fig:rd,tab:main} summarize our quantitative results across all 14 scenes. VersionGS achieves the highest mean PSNR and SSIM and the lowest LPIPS on both Technicolor and DanceNet3D, while remaining highly competitive with the top methods on N3DV at a fraction of their storage cost.

\paragraph{N3DV.}
The scenes of this widely used dataset are mostly static, and an unchanged Gaussian costs only a reference in VersionGS, so the dictionary stores 8--13\% as many primitives as independent per-frame reconstructions. Without pruning, VersionGS ranks third in PSNR, 0.23\,dB below MoRGS at $30\times$ less storage, and stays within 0.001 of the best SSIM and LPIPS. Our balanced setting requires the least storage in \cref{tab:main}, under half of StreamSTGS's, and has the second best temporal consistency. Since most baselines perform well on N3DV's slow motions, we focus our detailed analysis on Technicolor and DanceNet3D.

\paragraph{Technicolor: complex appearance and emerging content.}
At $\rho = 1$, VersionGS outperforms all baselines across PSNR, SSIM, and LPIPS while requiring $5.7\times$ less storage than TAD-GS. At $\rho = 0.35$, VersionGS uses the least storage of any method while matching or surpassing all baselines except TAD-GS in PSNR. With validated births adding new content (\cref{sec:ablation}), VersionGS cleanly handles objects appearing in the scenes. As the painter walks into the scene, our mean PSNR stays within 0.4\,dB of the empty scene. Crucially, when frames are ranked by motion magnitude (\cref{fig:motion}), VersionGS maintains flat reconstruction quality on the fastest frames, whereas most baselines drop by up to 3.1\,dB.

\input{fig/motion_quintiles}\paragraph{DanceNet3D: fast motion and deforming clothing.}
On DanceNet3D, rapid full-body motion causes most Gaussians to update every frame. Consequently, the GOP dictionary stores on average 80--95\% of the versions that independent frames would need, against 6--13\% on N3DV and Technicolor. By design, rather than smearing fast motion across time, VersionGS automatically allocates more versions to track the performer with exact per-frame states. VersionGS outperforms the strongest baseline by 2.7 dB in mean $\text{PSNR}_{\text{fg}}$ and leads across PSNR, SSIM, and LPIPS on every sequence. When ranking frames by inter-frame optical flow magnitude (\cref{fig:motion}), VersionGS loses less than 0.3 dB from the slowest quintile to the fastest, whereas baselines degrade by 1.1–4.0 dB. Even when trained like the baselines, with exact full image loss and black background, VersionGS still leads by 1.5\,dB on average, as the supplement shows.

\paragraph{Temporal consistency.}
A natural concern with per-frame Gaussian states is temporal flickering. However, as shown in \cref{tab:tlp}, VersionGS achieves competitive tLP on N3DV and the best temporal consistency on Technicolor and DanceNet3D, showing virtually no flicker within a GOP. The baselines score significantly worse on DanceNet3D because they do not reconstruct the performer's motion, as shown in \cref{fig:failures}.

\subsection{RD Trade-Off and Playback Speed}
\label{sec:rd}

We vary the dictionary keep fraction $\rho$ to control the rate--distortion (RD) trade-off. \Cref{fig:rd} compares the resulting curves with those of StreamSTGS. Moderate pruning preserves most reconstruction quality. Retaining half of the dictionary reduces mean PSNR by at most 0.19\,dB across all datasets. At $\rho=0.25$, N3DV and DanceNet3D remain within 0.33 and 0.04\,dB. Across matched bitrates, VersionGS outperforms StreamSTGS by up to 1.9 dB on N3DV, 3.9 dB on Technicolor, and 6.0 dB on DanceNet3D. On DanceNet3D, even our most aggressively pruned model surpasses every baseline in foreground PSNR.

\input{sec/tab_tlp}\paragraph{Decoding and playback.}
Because decoding a GOP only requires unpacking lossless WebP images and arithmetic-coded masks once per 60 frames, per-frame playback reduces to an index-gather operation followed by standard 3DGS rasterization. On a single NVIDIA H100 GPU, the time to render the very first frame of a sequence, including full 60-frame GOP decompression, averages at most 43 ms at $\rho = 0.35$ and 165 ms at $\rho = 1$. During continuous playback at $\rho = 0.35$, VersionGS achieves the lowest per-frame decoding overhead, and the highest frame rates across all datasets (\cref{tab:main}). Furthermore, integrating a lightweight decoder into the PlayCanvas viewer~\cite{playcanvas2026engine} enables real-time browser playback at 60 FPS on an iPhone 17 and 100 FPS on a MacBook Air M2 at $\rho = 0.35$, as the supplement shows.

\subsection{Where the Compression Comes From}
\label{sec:breakdown}

\Cref{tab:breakdown} breaks down how each stage of our pipeline contributes to compression.

\begin{itemize}[label={}, leftmargin=0pt]\setlength{\itemsep}{0pt}\setlength{\parskip}{0pt}
\item \textbf{Version sharing.} Reusing unchanged Gaussians across the 60-frame GOP reduces raw float32 storage by $9.3\times$ on flame steak and $14.1\times$ on painter without changing the Gaussian model, as the supplement shows. On big kicks, where the dancer moves continuously, sharing still saves 25\%.
\item \textbf{Image-based encoding.}
With the same codec and no adaptation, encoding each frame alone needs $10\times$ and $15\times$ more storage than our dictionary on flame steak and painter. Jointly encoding the concatenated frames~\cite{wang2026robust} lets the image codec remove much of this redundancy, yet still needs $3.1\times$, $4.4\times$ and $1.2\times$ more.
\item \textbf{Encoding-aware adaptation.} Joint GOP fine-tuning recovers quantization losses of up to 4.4\,dB while further shrinking WebP file sizes by 17--19\% via grid smoothing. However, on big kicks, where most versions serve a single frame, encoding alone loses no quality, and fine-tuning costs 0.3\,dB as attributes shift toward more compressible values.
\item \looseness=-1 \textbf{Pruning.} Compacting the dictionary to $\rho = 0.35$ reduces bitrate by a further 62--64\%. Weighting moving pixels with $\beta=32$ instead of $\beta=1$ changes mean PSNR by $-$0.08\,dB at the same bytes, lowers the mean loss of the worst tenth of frames, and increases the moving region's PSNR by up to 1\,dB (see supplement). Altogether, our pipeline compresses raw per-frame representations by $580\times$, $850\times$ and $74\times$.
\end{itemize}

\input{sec/tab_breakdown} 
\subsection{Construction Controls and Ablations}
\label{sec:ablation}

\looseness=-1 \Cref{tab:ablation} compares our construction with independent per-frame fitting and ablates its core design choices.

\begin{itemize}[label={}, leftmargin=0pt]\setlength{\itemsep}{0pt}\setlength{\parskip}{0pt}
\item \textbf{Independent frames.} Fitting every frame on its own needs $8.8\times$, $12\times$ and $1.3\times$ the storage at $\rho=1$.
\item \textbf{(i) Selective vs. unrestricted refinement.} Allowing gradient refinement to update all inherited Gaussians every frame inflates storage by $5.1\times$ on flame steak and $9.6\times$ on painter for a PSNR gain of at most 0.25\,dB. On big kicks, where most primitives move anyway, it still adds 22\% storage. Selective refinement is thus critical for preserving temporal reuse in static and partially moving scenes.
 \item \textbf{(ii) Motion initialization.} Disabling motion initialization lowers storage but degrades PSNR by 1.17\,dB on flame steak, 1.55\,dB on painter and 0.35\,dB on big kicks. While higher bitrates can easily be pruned later, tracking failures during sequential construction cannot be recovered during GOP adaptation.
\item \textbf{(iii) Validated births.}  In painter, the person is absent in frame 1 and walks into view. Without validated births, the pipeline is forced to stretch and recolor existing background Gaussians to cover the person, causing a 0.91 dB PSNR drop. Even though newly spawned primitives account for less than 1\% of the dictionary, they are essential for entering objects. 
\item \textbf{GOP length.} Increasing the GOP from $T=60$ to $T=120$ saves 10--20\% storage for at most 0.33\,dB, while $T=30$ costs 18--38\% more with no consistent quality gain.
\end{itemize}

\input{sec/tab_ablation}
\input{sec/5_conclusion}

%% file: fig/motion_quintiles.tex
\begin{figure}[t]
  \centering
  \begin{tikzpicture}\begin{axis}[width=0.535\columnwidth, height=4.1cm, xmin=0.7, xmax=5.3, xtick={1,2,3,4,5}, xticklabels={low,,,,high}, ymin=29, ymax=34.5, grid=major, grid style={gray!25}, title={Technicolor}, title style={font=\small\bfseries, yshift=-5pt}, xlabel={frame motion}, xlabel style={font=\scriptsize, yshift=4pt}, ylabel={PSNR (dB)}, ylabel style={font=\scriptsize, yshift=-5pt}, tick label style={font=\tiny}]
    \addplot[line width=0.7pt, mark size=1.8pt, color=rdC1, mark=square*] coordinates {(1,33.28) (2,33.36) (3,33.27) (4,33.05) (5,32.70)};
    \addplot[line width=0.7pt, mark size=1.8pt, color=rdC2, mark=diamond*, mark size=2.2pt] coordinates {(1,30.14) (2,30.18) (3,30.30) (4,30.33) (5,30.14)};
    \addplot[line width=0.7pt, mark size=1.8pt, color=rdC3, mark=triangle*, mark size=2.2pt] coordinates {(1,32.88) (2,31.44) (3,30.90) (4,30.99) (5,29.80)};
    \addplot[line width=0.7pt, mark size=1.8pt, color=rdC6, mark=*] coordinates {(1,32.83) (2,32.39) (3,32.07) (4,31.43) (5,30.28)};
    \addplot[line width=0.7pt, mark size=1.8pt, color=rdC8, mark=oplus*] coordinates {(1,32.09) (2,31.05) (3,30.54) (4,30.37) (5,29.24)};
    \addplot[line width=0.7pt, mark size=1.8pt, color=rdC7, mark=otimes*] coordinates {(1,34.27) (2,34.29) (3,34.20) (4,33.42) (5,31.72)};
    \addplot[line width=1.1pt, mark size=1.8pt, color=rdOurs, mark=star, mark size=2.8pt, line width=1.0pt] coordinates {(1,33.03) (2,33.14) (3,33.39) (4,33.09) (5,33.01)};
  \end{axis}\end{tikzpicture}\hfill
  \begin{tikzpicture}\begin{axis}[width=0.535\columnwidth, height=4.1cm, xmin=0.7, xmax=5.3, xtick={1,2,3,4,5}, xticklabels={low,,,,high}, ymin=16, ymax=26.5, grid=major, grid style={gray!25}, title={DanceNet3D}, title style={font=\small\bfseries, yshift=-5pt}, xlabel={frame motion}, xlabel style={font=\scriptsize, yshift=4pt}, ylabel={PSNR$_\text{fg}$ (dB)}, ylabel style={font=\scriptsize, yshift=-5pt}, tick label style={font=\tiny}, legend to name=motionlegend, legend columns=3, legend style={font=\scriptsize, draw=none, column sep=3pt}]
    \addplot[line width=0.7pt, mark size=1.8pt, color=rdC1, mark=square*] coordinates {(1,20.14) (2,18.97) (3,18.35) (4,17.21) (5,16.86)};
    \addlegendentry{Ex4DGS}
    \addplot[line width=0.7pt, mark size=1.8pt, color=rdC2, mark=diamond*, mark size=2.2pt] coordinates {(1,22.92) (2,22.60) (3,22.45) (4,22.19) (5,21.84)};
    \addlegendentry{4DGS}
    \addplot[line width=0.7pt, mark size=1.8pt, color=rdC3, mark=triangle*, mark size=2.2pt] coordinates {(1,22.94) (2,21.19) (3,20.53) (4,19.74) (5,18.92)};
    \addlegendentry{MoRGS}
    \addplot[line width=0.7pt, mark size=1.8pt, color=rdC6, mark=*] coordinates {(1,22.46) (2,21.55) (3,21.09) (4,20.49) (5,20.24)};
    \addlegendentry{StreamSTGS}
    \addplot[line width=0.7pt, mark size=1.8pt, color=rdC8, mark=oplus*] coordinates {(1,24.37) (2,23.49) (3,23.16) (4,22.39) (5,21.21)};
    \addlegendentry{QUEEN}
    \addplot[line width=0.7pt, mark size=1.8pt, color=rdC7, mark=otimes*] coordinates {(1,23.63) (2,22.90) (3,22.52) (4,21.80) (5,20.92)};
    \addlegendentry{TAD-GS}
    \addplot[line width=1.1pt, mark size=1.8pt, color=rdOurs, mark=star, mark size=2.8pt, line width=1.0pt] coordinates {(1,25.78) (2,25.74) (3,25.75) (4,25.61) (5,25.61)};
    \addlegendentry{VersionGS (ours)}
  \end{axis}\end{tikzpicture}%
  \\[-2pt]
  \ref{motionlegend}
  \caption{\textbf{Quality against motion.} Held-out frames of each sequence are split into fifths by the mean optical flow between consecutive training photographs, in the person mask on DanceNet3D, and PSNR is averaged per fifth. Ours at $\rho=0.35$. Our quality holds as motion grows while most baselines drop.}
  \label{fig:motion}
\end{figure}
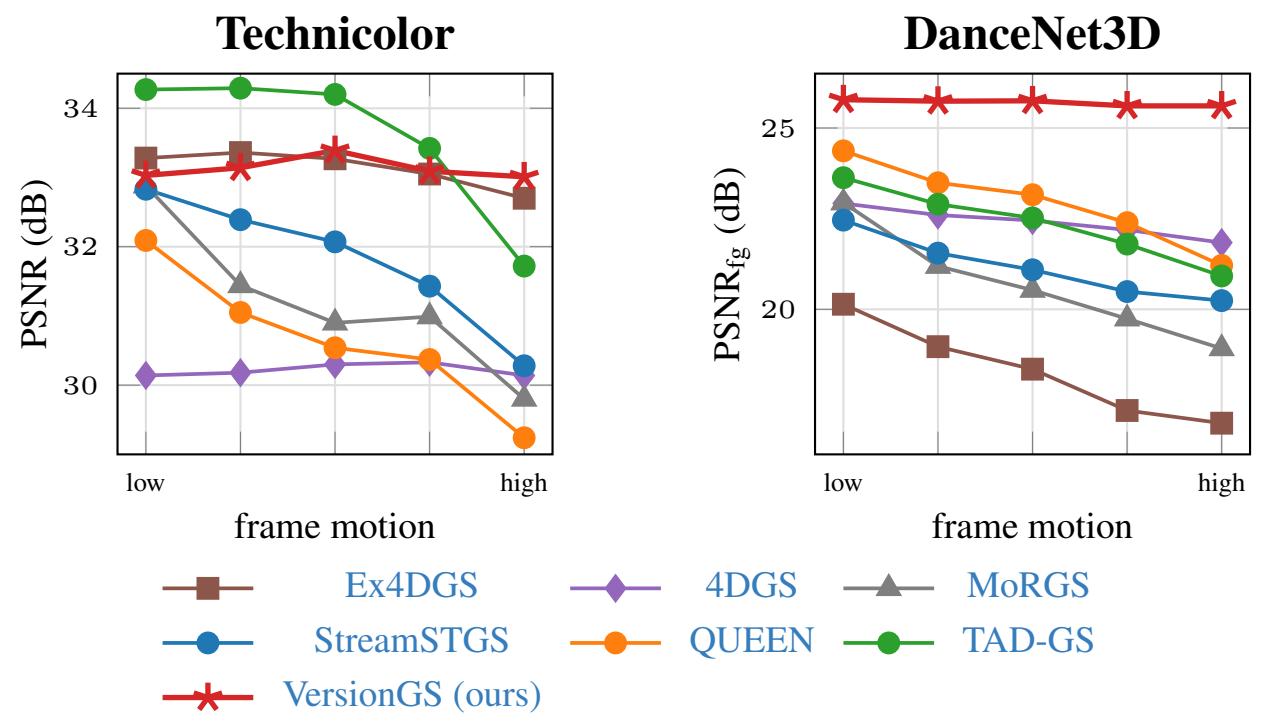

%% file: sec/tab_tlp.tex
\begin{table}[t]
  \centering
  \setlength{\tabcolsep}{5pt}
  \resizebox{\columnwidth}{!}{%
  \begin{tabular}{l ccc}
    \toprule
    Method & N3DV & Technicolor & DanceNet3D \\
    \midrule
    Ex4DGS~\cite{lee2024ex4dgs}           & 1.39 & 0.40 & 6.99 \\
    4DGS~\cite{yang2023gs4d}              & 1.38 & 0.77 & 2.74 \\
    3DGStream~\cite{sun20243dgstream}     & \rankA{0.94} & 0.56 & 4.12 \\
    MoRGS~\cite{lee2026morgs}           & 1.28 & 0.47 & 3.95 \\
    D-FCGS~\cite{zhang2025dfcgs}          & 1.21 & 0.81 & 6.55 \\
    StreamSTGS~\cite{ke2026streamstgs}    & 1.38 & 0.50 & 4.75 \\
    QUEEN~\cite{girish2024queen}          & \rankC{1.07} & \rankC{0.31} & \rankC{1.80} \\
    TAD-GS~\cite{sandu2026tadgs}          & 1.37 & 0.54 & 4.82 \\
    \midrule
    VersionGS (ours, no pruning)        & 1.17 & \rankA{0.23} & \rankA{0.67} \\
    VersionGS (ours, balanced RD)         & \rankB{1.03} & \rankB{0.27} & \rankB{0.69} \\
    \bottomrule
  \end{tabular}}
  \caption{\textbf{Temporal consistency.} tLP$\times100$ on the held-out cameras, on DanceNet3D in the person's bounding box. \textbf{Lower is better}.}
  \label{tab:tlp}
\end{table}

%% file: sec/tab_breakdown.tex
\begin{table}[t]
  \centering
  \setlength{\tabcolsep}{2.6pt}
  \resizebox{\columnwidth}{!}{%
  \begin{tabular}{l cc cc cc}
    \toprule
    & \multicolumn{2}{c}{flame steak} & \multicolumn{2}{c}{painter} & \multicolumn{2}{c}{big kicks} \\
    \cmidrule(lr){2-3}\cmidrule(lr){4-5}\cmidrule(lr){6-7}
    Stage & MB/fr & PSNR & MB/fr & PSNR & MB/fr & PSNR$_\text{fg}$ \\
    \midrule
    Each frame, float32                 & 35.4 & 33.80 & 71.3 & 36.58 & 11.8 & 26.03 \\
    Each frame encoded alone            & 2.01 & 33.08 & 3.97 & 34.03 & 0.734 & 26.05 \\
    Frames joined~\cite{wang2026robust} & 0.624 & 33.68 & 1.16 & 33.81 & 0.643 & 26.01 \\
    \midrule
    Version dictionary, encoded         & 0.199 & 33.59 & 0.265 & 32.41 & 0.541 & 26.01 \\
    Adaptation ($\rho=1$)               & 0.161 & 34.16 & 0.218 & 36.85 & 0.449 & 25.72 \\
    Pruning, $\beta=1$ ($\rho=0.35$)    & 0.061 & 33.95 & 0.084 & 36.69 & 0.161 & 25.73 \\
    Pruning, $\beta=32$ ($\rho=0.35$)   & 0.061 & 33.92 & 0.084 & 36.46 & 0.160 & 25.74 \\
    \bottomrule
  \end{tabular}}
  \caption{\textbf{Where the compression comes from}, on the first GOP of flame steak, painter and big kicks. Rows two and three code the same constructed frames with the same image codec and no adaptation. The other rows follow our pipeline, and $\beta$ weights moving pixels in the pruning score of \cref{eq:prune_motion}.}
  \label{tab:breakdown}
\end{table}

%% file: sec/tab_ablation.tex
\begin{table}[t]
  \centering
  \setlength{\tabcolsep}{2.6pt}
  \resizebox{\columnwidth}{!}{%
  \begin{tabular}{l cc cc cc}
    \toprule
    & \multicolumn{2}{c}{flame steak} & \multicolumn{2}{c}{painter} & \multicolumn{2}{c}{big kicks} \\
    \cmidrule(lr){2-3}\cmidrule(lr){4-5}\cmidrule(lr){6-7}
    Variant & MB/fr & PSNR & MB/fr & PSNR & MB/fr & PSNR$_\text{fg}$ \\
    \midrule
    Full method ($\rho=1$)              & 0.161 & 34.16 & 0.218 & 36.85 & 0.449 & 25.72 \\
    Independent frames ($\rho=1$)       & 1.424 & 33.92 & 2.691$^\dagger$ & 36.47 & 0.583 & 25.60 \\
    (i) unrestricted refinement         & 0.817 & 34.34 & 2.100 & 37.10 & 0.546 & 25.70 \\
    (ii) w/o motion initialization      & 0.110 & 32.99 & 0.152 & 35.30 & 0.181 & 25.37 \\
    (iii) w/o validated births          & 0.164 & 33.66 & 0.209 & 35.94 & 0.449 & 25.67 \\
    \midrule
    Full method ($\rho=0.35$)           & 0.061 & 33.92 & 0.084 & 36.46 & 0.160 & 25.74 \\
    Independent, matched rate   & 0.062 & 27.93 & 0.079 & 28.53 & 0.157 & 25.60 \\
    \bottomrule
  \end{tabular}}
  \caption{\textbf{Construction controls and ablation} on the first GOP of the scenes of \cref{tab:breakdown}. Independent frames fits every frame on its own and then runs our pipeline. $^\dagger$At $\rho=0.9$.}
  \label{tab:ablation}
\end{table}

%% file: sec/5_conclusion.tex
\section{Discussion and Limitations}
\label{sec:limitations}

While VersionGS achieves strong rate-distortion performance and motion robustness, several limitations remain for future work. Because VersionGS stores discrete per-frame states without continuous temporal interpolation, subtle high-frequency flicker can occasionally appear on fast-moving surfaces, and visible transitions can occur across independent GOP boundaries. 
Sequential frame-by-frame construction reuses versions locally across consecutive frames. If an object moves away and later returns to an earlier pose, new versions will be created rather than matching against distant historical states. Also, overall GOP quality remains sensitive to the initial first-frame MCMC reconstruction. 
Building a 60-frame GOP takes 3–8 GPU-hours, and playback of a GOP begins only after its dictionary images are decoded.
Lossy WebP saved about 15\% of the bytes at a scene-dependent cost in quality, as the supplement shows. Storing each version as a residual of its predecessor lowered the rate by 17\% on average, but decoding took longer with 2.2$\times$ the peak memory, and chaining versions raised the PSNR loss of lossy coding from at most 1.3\,dB to up to 4.4\,dB.

\section{Conclusion}
\label{sec:conclusion}

We presented VersionGS, a dynamic Gaussian splatting representation that replaces fragile temporal motion models with a shared dictionary of explicit, versioned Gaussian states. By updating only primitives with clear evidence of change, selective sequential construction preserves exact temporal reuse in static regions while decoupling visual quality from motion speed. It achieves top tier reconstruction quality on Technicolor and DanceNet3D, matches the top methods on N3DV at a fraction of the bitrate, and scales flexibly across bitrates via GOP-level pruning without rebuilding. Finally, because rendering requires only a binary mask lookup and a standard 3DGS rasterizer, VersionGS brings high-fidelity, real-time free-viewpoint video directly to everyday web browsers on laptops and mobile phones. More broadly, our results show that explicit state deduplication paired with static image coding offers a simpler, more resilient foundation for streamable 4D video than continuous motion parameterization.

%% file: sec/llm_statement.tex
\clearpage %
\section*{LLM Usage Statement}
We used Claude Code. It wrote much of the code for the method, experiments, and baselines comparison, helped run experiments, drafted text and checked the references. The authors designed the method and experiments, completed the text, verified all results and references, and take full responsibility for the paper.

%% file: sec/X_suppl.tex
\clearpage
\setcounter{page}{1}
\maketitlesupplementary

\section{Supplementary Video}
\label{sec:supp_video}

The supplementary material contains a video, \texttt{supplementary\_video.mp4}. It shows all 14 sequences of \cref{tab:main} on a held-out camera, in the order DanceNet3D, Technicolor, N3DV. For each sequence, it plays all evaluated frames at 30 frames per second, with the held-out photograph, our method at $\rho = 1$, and four baselines side by side. The baselines are QUEEN, StreamSTGS, TAD-GS, and MoRGS on N3DV and DanceNet3D, and TAD-GS, Ex4DGS, StreamSTGS, and QUEEN on Technicolor. All images are the renders that \cref{tab:main} scores, and each one carries its PSNR on the current frame, within the person mask on DanceNet3D. The video ends with three segments that cross a GOP boundary at quarter speed, on big kicks, painter, and cut roasted beef, with QUEEN, StreamSTGS, and a row of frame-to-frame differences. Each segment shows the boundary at which the tLP of our method is largest on that camera.

On N3DV, our renders are on par with those of the best baselines. On Technicolor and DanceNet3D, they are visibly better where the content moves. On DanceNet3D, our method reconstructs the fast-moving dancers sharply and accurately. On Technicolor, it handles occlusion and motion cleanly. The transition at a GOP boundary is visible in a few daytime N3DV scenes and is not visible in most other scenes. From the first ten to the last ten frames of a GOP, the PSNR on the held-out cameras changes by $+0.1$\,dB on N3DV, $-0.3$\,dB on Technicolor, and $-0.8$\,dB on DanceNet3D on average.

\section{DanceNet3D Trained with Baseline Loss Settings}
\label{sec:supp_armE}

On DanceNet3D, the baselines are trained on foreground images composited over a black background using a standard full-frame loss, except StreamSTGS on \emph{biz markie}, which failed on black and is trained on white because the performer wears mostly black and 57\% of the person pixels have a luminance below 0.08. In contrast, our default configuration normalizes the $L_1$ loss by the foreground mask area, and samples random background colors during optimization. To isolate the impact of our representation from these training choices, we re-run VersionGS on all five GOPs of each sequence using a fixed black background and standard full-image $L_1$ and SSIM losses across all stages, including first-frame initialization, sequential construction, and GOP adaptation, while keeping all other hyperparameters unchanged. We evaluate the unpruned models on the same frames and held-out views using the protocol from \cref{tab:main}.

As shown in \cref{tab:supp_armE}, adopting the baseline training loss reduces our PSNR$_\text{fg}$ by 0.5--1.9\,dB compared to our default configuration, yet VersionGS still outperforms the strongest baseline on every sequence by 0.8--2.0\,dB while achieving lower LPIPS$_\text{fg}$. The training objective accounts for 0.5--1.9\,dB of our result, and a lead of 0.8--2.0\,dB remains under the baselines' own objective. The lead in perceptual quality is far larger and holds under both objectives. Averaged over the four sequences, our LPIPS$_\text{fg}$ is 44\% below that of the best baseline with our default loss and still 26\% below it with the baselines' loss, and it is lower on every sequence in both cases.

\begin{table}[h]
  \centering
  \small
  \setlength{\tabcolsep}{3pt}
  \resizebox{\columnwidth}{!}{%
  \begin{tabular}{l ccc ccc}
    \toprule
    & \multicolumn{3}{c}{PSNR$_\text{fg}\uparrow$} & \multicolumn{3}{c}{LPIPS$_\text{fg}\downarrow$} \\
    \cmidrule(lr){2-4}\cmidrule(lr){5-7}
    Sequence & Baseline loss & Ours & Best baseline & Baseline loss & Ours & Best baseline \\
    \midrule
    chimee         & 24.26 & 25.35 & 22.29 & 0.197 & 0.147 & 0.267 \\
    russia costume & 21.17 & 22.38 & 20.39 & 0.192 & 0.151 & 0.290 \\
    big kicks      & 24.55 & 25.08 & 22.54 & 0.128 & 0.104 & 0.167 \\
    biz markie     & 27.96 & 29.90 & 26.76 & 0.197 & 0.136 & 0.242 \\
    \bottomrule
  \end{tabular}}
  \caption{Comparison of VersionGS trained with the baseline loss configuration against our default configuration. Baseline loss uses a fixed black background and full-frame objective. Both are evaluated on all 300 frames of each DanceNet3D sequence at $\rho = 1$. Best baseline reports the highest-scoring baseline in PSNR$_\text{fg}$ on the same frames.}
  \label{tab:supp_armE}
\end{table}

\section{\texorpdfstring{Versioned Dictionary vs.\ Joint Static Coding}{Versioned Dictionary vs. Joint Static Coding}}
\label{sec:supp_versioning}

\leavevmode To isolate the compression gains of explicit versioning from those of the image codec, \cref{tab:supp_versioning} evaluates different ways of encoding the same 60 frames from the first GOP using our image codec without adaptation, following the protocol of \cref{tab:breakdown}. Following Wang \etal~\cite{wang2026robust}, the ``Joined, one set'' variant concatenates the Gaussians of all 60 frames into a single pool prior to PLAS sorting. Thus, a Gaussian that remains unchanged across 40 frames is stored as 40 identical copies that the sorter places in adjacent grid cells. Deduplicating these exact copies reduces the joined pool to our versioned dictionary. The SH codebook has about $K/32$ entries and at most $256^2$, so in \cref{tab:breakdown} the joined pool, which holds more records, also gets a larger codebook than our dictionary. With the same $256^2$ codebook, explicit deduplication via our dictionary reduces storage by $2.9\times$, $3.8\times$, and $1.2\times$ at virtually identical PSNR.

\leavevmode The lower block of \cref{tab:supp_versioning} applies the same joint coding to frames reconstructed independently using our first-frame MCMC configuration. Without inter-frame tracking, joint coding saves only 8\% on flame steak, 8\% on painter, and under 1\% on big kicks, confirming Wang \etal's observation that joint static coding is effective only when primitives are tracked across time. Thus, the redundancy exploited by our dictionary arises directly from our selective sequential construction, which preserves exact states for static primitives. Even on big kicks, where nearly every Gaussian moves, independent per-frame reconstruction requires $1.3\times$ the storage of our dictionary while scoring 0.2\,dB lower in PSNR$_\text{fg}$.

\leavevmode \Cref{tab:supp_indep} further compares independent per-frame reconstructions against our sequential frames across the full pipeline, including pruning, encoding, and adaptation, as in \cref{tab:ablation}. Without pruning, independent frames require $8.8\times$, $12\times$, and $1.3\times$ the storage of VersionGS. When pruned to match the bitrate of our balanced setting, $\rho = 0.35$, independent reconstructions retain only 3.5\%, 2.2\%, and 25\% of their Gaussians, suffering PSNR drops of 6.0, 7.9, and 0.0\,dB while LPIPS degrades to 0.223, 0.258, and 0.151. On the first GOP of the other three DanceNet3D sequences, at matched storage, our method is 1.2\,dB above independent frames on russia costume, 0.1\,dB above on biz markie, and 0.5\,dB below on chimee. In this low-rate setting, independent frames flicker, and our method largely eliminates this flicker. On the first GOP of chimee, the tLP of independent frames in the person box is 0.9 times ours, and on big kicks, russia costume, and biz markie, it is 1.8 to 3.7 times ours at matched storage.

\begin{table}[h]
  \centering
  \small
  \setlength{\tabcolsep}{3pt}
  \resizebox{\columnwidth}{!}{%
  \begin{tabular}{l cc cc cc}
    \toprule
    & \multicolumn{2}{c}{flame steak} & \multicolumn{2}{c}{painter} & \multicolumn{2}{c}{big kicks} \\
    \cmidrule(lr){2-3}\cmidrule(lr){4-5}\cmidrule(lr){6-7}
    Coding & MB/fr & PSNR & MB/fr & PSNR & MB/fr & PSNR$_\text{fg}$ \\
    \midrule
    \multicolumn{7}{l}{\emph{raw float32 storage}} \\
    Per-frame copies                & 35.4 & 33.80 & 71.3  & 36.58 & 11.8  & 26.03 \\
    Version dictionary              & 3.79 & 33.80 & 5.04 & 36.58 & 8.83 & 26.03 \\
    \midrule
    \multicolumn{7}{l}{\emph{our frames}} \\
    Dictionary, as in \cref{tab:breakdown} & 0.199 & 33.59 & 0.265 & 32.41 & 0.541 & 26.01 \\
    Dictionary, $256^2$ codebook  & 0.215 & 33.62 & 0.309 & 33.80 & 0.541 & 26.01 \\
    Joined, one set                 & 0.624 & 33.68 & 1.162 & 33.81 & 0.643 & 26.01 \\
    Each frame alone                & 2.010 & 33.08 & 3.968 & 34.03 & 0.734 & 26.05 \\
    \midrule
    \multicolumn{7}{l}{\emph{frames reconstructed independently}} \\
    Joined, one set                 & 1.891 & 32.84 & 3.697 & 33.31 & 0.700 & 25.86 \\
    Each frame alone                & 2.059 & 32.99 & 4.013 & 33.76 & 0.703 & 25.85 \\
    \bottomrule
  \end{tabular}}
  \caption{\textbf{Versions against joint static coding} on the first GOP, same image codec, no adaptation. Painter is coded in two spatial slices, due to the size limit of the codec.}
  \label{tab:supp_versioning}
\end{table}
\input{sec/tab_indep}
\section{Decoding and Playback}
\label{sec:supp_decode}
\input{sec/tab_decode}
\input{sec/tab_browser}\Cref{tab:decode} benchmarks decoding and rendering latencies across all 70 GOPs from \cref{tab:main} using each method's native decoder and rasterizer, starting from the files that \cref{tab:main} counts. All measurements are conducted on a single NVIDIA H100 80GB GPU and 14 cores of an Intel Xeon Platinum 8480+, reporting the median of three passes after a warm-up pass and initial file read. Decode measures the total GOP decoding time divided by the number of frames required to reconstruct the Gaussian primitives for the rasterizer. Playback adds the time to render a single held-out view. First frame measures the cold-start latency from the beginning of decoding until the first frame is rendered. For VersionGS, this includes decompressing the entire first GOP dictionary, gathering the active primitives for frame 1, and rasterizing.

At $\rho = 0.35$, VersionGS requires under 1.2\,ms per frame to decode across all datasets, even on the slowest GOP, achieving the lowest decoding and playback latencies of all evaluated methods. While QUEEN, MoRGS, and 3DGStream achieve lower first-frame latency on DanceNet3D because they store the initial frame directly without GOP-level bundling, their subsequent frames require roughly 200\,ms, 250\,ms, and 6\,ms per frame to decode, respectively. Because each GOP in VersionGS is self-contained, random access to any GOP boundary takes at most 62\,ms at $\rho = 0.35$ and 202\,ms at $\rho = 1$.

\textbf{Playback speed.} The FPS column in \cref{tab:main} is end-to-end playback throughput: the number of frames played, all 300, 290 for \emph{birthday}, divided by the wall time of the pass. The clock starts once libraries are loaded, the CUDA context is warmed with one dummy render and every stored file has been read into memory once, so disk reads are excluded, while parsing and decompressing the stored files are included. Each method then processes the frames in display order without pre-decoding: it decodes the frame from the files that \cref{tab:main} counts with its own released decoder, including entropy decoding, image or video decoding and any network evaluation, renders the held-out camera once with its own rasterizer at its evaluation resolution, and synchronizes the GPU before the next frame. Display, vertical sync and prefetching of later GOPs are excluded. We report the median of three passes after an untimed warm-up pass, on the hardware of \cref{tab:decode}. Decoding a frame consists of:
\begin{itemize}
\setlength{\itemsep}{0pt}
\item VersionGS: when playback reaches a GOP, decoding its dictionary images, codebooks and frame masks on 14 CPU threads, then gathering each frame's active versions. No network.
\item Ex4DGS: interpolating every dynamic Gaussian between keyframes, with cubic positions, spherical rotations and temporal opacity. No network.
\item 4DGS: slicing every 4D Gaussian at time $t$ into a conditional 3D mean, covariance and temporal opacity, which its released renderer does inside the rasterizer. No network.
\item TAD-GS: evaluating every Gaussian's Fourier motion, rotation and opacity at time $t$. No network.
\item 3DGStream: evaluating the frame's neural transformation cache, a hash grid with an MLP, on all Gaussians and appending the frame's new Gaussians.
\item D-FCGS: decoding each P frame with the released feed-forward decoder network and its zstd-coded positions; I frames are loaded from their PLY files.
\item StreamSTGS: decoding each GOP's H.265 feature video and JPEG XL planes, then evaluating its temporal and attribute MLPs for the held-out camera in every frame.
\item QUEEN and MoRGS: entropy decoding each frame's quantized latents, decoding them into attribute residuals with the frame's learned decoders and applying them to the previous frame's Gaussians, which the released code runs largely on the CPU. Once decoded, a frame renders in 2 to 3\,ms on N3DV and Technicolor, consistent with the real-time rendering both papers report, so their FPS is limited by decoding at 1.2 to 1.5\,s per frame.
\end{itemize}
At $\rho = 0.35$, VersionGS achieves the fastest playback on every dataset, ranging from 286 to 632 FPS across scenes.

To demonstrate lightweight client-side execution, we integrated a JavaScript and WebAssembly decoder into the PlayCanvas SuperSplat web viewer~\cite{playcanvas2026engine}, enabling browser playback on mobile phones and laptops. \Cref{tab:browser} reports full-resolution Chrome benchmarks on an iPhone 17 and a MacBook Air M2 across one representative scene per dataset. Decompressing a 60-frame GOP on the iPhone takes at most 324\,ms, well below the 2-second playback duration of a 30 FPS GOP, allowing subsequent GOPs to be prefetched seamlessly in the background. VersionGS achieves real-time playback on all scenes, exceeding 60 FPS on the iPhone 17 and 100 FPS on the MacBook Air M2 at $\rho = 0.35$. In the normal player at 30 frames per second, decoding never holds up playback on any scene.

\section{Motion Weighting in Dictionary Pruning}
\label{sec:supp_beta}
\input{sec/tab_beta}\leavevmode \Cref{tab:beta} evaluates the impact of the motion weight $\beta$ in the pruning score of \cref{eq:prune_motion} on the first GOP of the three scenes from \cref{tab:breakdown} at $\rho = 0.35$. At matched bitrates, setting $\beta = 32$ improves PSNR on moving pixels by up to 1.0\,dB over uniform weighting, $\beta = 1$, and reduces quality degradation on the worst 10\% of frames on flame steak and painter. On big kicks, where the foreground performer dominates the changing pixels, performance remains stable within 0.02\,dB across all non-random $\beta$ values. Increasing the weight further to $\beta = 128$ over-penalizes static background primitives and lowers full-frame PSNR by up to 0.3\,dB, whereas random pruning lowers PSNR by up to 1.7\,dB against $\beta=32$ and has a higher LPIPS on every scene. Averaged over the three scenes, $\beta=32$ gives the lowest worst-10\% loss, 0.50\,dB against 0.60\,dB for $\beta=1$ and $\beta=8$, and a moving-region PSNR within 0.01\,dB of the best, at 0.08\,dB below $\beta=1$ in mean PSNR and with equal LPIPS.

\section{Lossy Coding of Attribute Images}
\label{sec:supp_lossy}
\input{sec/tab_lossy}\leavevmode \Cref{tab:supp_lossy} compares lossy WebP compression of selected attribute images against our default lossless encoder under identical encoding-aware adaptation. Applying lossy compression to the rotation grids and the lower byte of the 16-bit center coordinates reduces total storage by 14--15\% across all datasets and keep fractions $\rho$, while shifting mean PSNR by at most 0.21\,dB in the top half of \cref{tab:supp_lossy}.

The bottom half of \cref{tab:supp_lossy} evaluates lossy coding across additional attribute grids at $\rho = 0.35$ on the three scenes from \cref{tab:breakdown}. Compressing any single attribute grid saves about 8\% of the total bitrate. Combining lossy rotations, scales, and center low bytes, or reducing the WebP quality of the center low byte to 75, increases storage savings to 18--21\%. While PSNR remains within 0.25\,dB of the lossless baseline on flame steak and within 0.04\,dB on big kicks, quality on painter fluctuates between $-0.75$\,dB and $+0.51$\,dB without a monotonic trend. Because lossy attribute coding introduces scene-dependent quality variations, we use lossless WebP for all main results.

\section{Versions Stored as Residuals}
\label{sec:supp_residual}
\input{sec/tab_residual}\leavevmode \Cref{tab:supp_residual} gives the measurements behind the residual experiment in the conclusion of the main paper. Both forms store the same versions of the first GOP. In the residual form, the first version of a Gaussian is stored as in our method, and each later version is stored as its difference from the previous version of the same Gaussian. Both forms use the same adaptation and lossless coding, and both are decoded on one H100 GPU with 14 CPU threads.

The residual form is 14--22\% smaller in \cref{tab:supp_residual}, and 17\% smaller on average over 22 rate points of seven GOPs, at a PSNR within 0.1\,dB except on painter at $\rho=1$. It costs little decoding time, 1.22$\times$ on average, and it reaches the first frame sooner, because frame 1 needs no residuals. Its costs are elsewhere. Decoding holds 2.1--2.4$\times$ the peak GPU memory. A version can be recovered only through its chain of up to 59 predecessors, so one corrupted record makes 8--25 records wrong on average, against one for absolute versions. With lossy WebP on the rotation grids and the low center byte, errors add up along the chain, and the PSNR loss on flame steak and painter grows from 1.1--1.3\,dB to 3.1--4.4\,dB. We keep absolute versions for this reason and leave the residual form as an option for lower rates.

%% file: sec/tab_indep.tex
\begin{table*}[t]
  \centering
  \small
  \setlength{\tabcolsep}{4pt}
  \resizebox{\ifdim\width>\textwidth \textwidth\else\width\fi}{!}{%
  \begin{tabular}{l cccc cccc cccc}
    \toprule
    & \multicolumn{4}{c}{flame steak} & \multicolumn{4}{c}{painter} & \multicolumn{4}{c}{big kicks} \\
    \cmidrule(lr){2-5}\cmidrule(lr){6-9}\cmidrule(lr){10-13}
    Frames & $\rho$ & MB/fr & PSNR & LPIPS & $\rho$ & MB/fr & PSNR & LPIPS & $\rho$ & MB/fr & PSNR$_\text{fg}$ & LPIPS \\
    \midrule
    Ours, no pruning & 1 & 0.161 & 34.16 & 0.062 & 1 & 0.218 & 36.85 & 0.038 & 1 & 0.449 & 25.72 & 0.125 \\
    Independent, no pruning & 1 & 1.424 & 33.92 & 0.064 & 0.9$^\dagger$ & 2.691 & 36.47 & 0.033 & 1 & 0.583 & 25.60 & 0.127 \\
    \midrule
    Ours, balanced & 0.35 & 0.061 & 33.92 & 0.071 & 0.35 & 0.084 & 36.46 & 0.048 & 0.35 & 0.160 & 25.74 & 0.137 \\
    Independent, at the bytes of the row above & 0.035 & 0.062 & 27.93 & 0.223 & 0.022 & 0.079 & 28.53 & 0.258 & 0.25 & 0.157 & 25.60 & 0.151 \\
    \bottomrule
  \end{tabular}}
  \caption{\textbf{Independent per-frame fits through our pipeline} (pruning, encoding and adaptation) against our method, on the first GOP. $\rho$ is the kept fraction of each row's own Gaussians. Before coding, the independent fits reach 33.65, 35.40 and 25.80\,dB, against 33.80, 36.58 and 26.03\,dB for our frames. $^\dagger$At $\rho=1$ painter's 18\,M Gaussians exceed one image grid, and $\rho=0.9$ removes only Gaussians with a zero pruning score.}
  \label{tab:supp_indep}
\end{table*}

%% file: sec/tab_decode.tex
\begin{table*}[t]
  \centering
  \setlength{\tabcolsep}{4pt}
  \resizebox{\textwidth}{!}{%
  \begin{tabular}{l ccc ccc ccc}
    \toprule
    & \multicolumn{3}{c}{N3DV (6 scenes)} & \multicolumn{3}{c}{Technicolor (4 scenes)} & \multicolumn{3}{c}{DanceNet3D (4 scenes)} \\
    \cmidrule(lr){2-4}\cmidrule(lr){5-7}\cmidrule(lr){8-10}
    Method & Decode$\downarrow$ & Playback$\downarrow$ & First frame$\downarrow$
           & Decode$\downarrow$ & Playback$\downarrow$ & First frame$\downarrow$
           & Decode$\downarrow$ & Playback$\downarrow$ & First frame$\downarrow$ \\
    \midrule
        Ex4DGS~\cite{lee2024ex4dgs}          & \rankC{2.45} & 12.0 & 541 & 11.2 & 15.8 & 490 & \rankB{0.97} & \rankB{1.91} & 121 \\
        4DGS~\cite{yang2023gs4d}             & --$^\dagger$ & \rankC{6.78$^\dagger$} & 364 & --$^\dagger$ & 14.4$^\dagger$ & 2234 & --$^\dagger$ & \rankC{3.61$^\dagger$} & 379 \\
        3DGStream~\cite{sun20243dgstream}  & 5.85 & 8.80 & \rankC{131} & \rankC{6.98} & \rankC{10.1} & 352 & 5.54 & 6.87 & \rankC{32} \\
        MoRGS~\cite{lee2026morgs}          & 1468 & 1472 & 170 & 1522 & 1525 & \rankC{178} & 246 & 248 & \rankB{22} \\
        D-FCGS~\cite{zhang2025dfcgs}       & 186 & 191 & 143 & 629 & 634 & 354 & 52.3 & 54.9 & 71 \\
        StreamSTGS~\cite{ke2026streamstgs}   & 37.2 & 40.2 & 2181 & 39.4 & 40.6 & 2203 & 17.0 & 17.9 & 1273 \\
        QUEEN~\cite{girish2024queen}         & 1315 & 1318 & 157 & 1203 & 1205 & 187 & 202 & 204 & \rankA{14} \\
        TAD-GS~\cite{sandu2026tadgs}         & 5.85 & 10.0 & 1449 & 10.2 & 13.3 & 2636 & 4.47 & 8.22 & 1064 \\
    \midrule
        VersionGS (ours, no pruning)         & \rankB{1.41} & \rankB{4.90} & \rankB{77} & \rankB{2.02} & \rankB{4.32} & \rankB{82} & \rankC{2.95} & 3.96 & 165 \\
        VersionGS (ours, balanced RD)        & \rankA{0.47} & \rankA{3.26} & \rankA{24} & \rankA{0.68} & \rankA{2.37} & \rankA{29} & \rankA{0.85} & \rankA{1.71} & 43 \\
    \bottomrule
  \end{tabular}}
  \caption{\textbf{Decoding and playback} in milliseconds on one NVIDIA H100 GPU. Decode is the total time to decode all frames of a GOP divided by the frame count, and playback is decode plus the time to render one held-out view, both per frame and averaged over the GOPs of each dataset. First frame is the time until the first frame of the sequence is rendered once, including all decoding it needs, averaged over the scenes. \protect\colorbox{red!25}{Best}, \protect\colorbox{orange!25}{second} and \protect\colorbox{yellow!35}{third} per column. $^\dagger$4DGS slices its Gaussians inside its rasterizer.}
  \label{tab:decode}
\end{table*}

%% file: sec/tab_browser.tex
\begin{table}[t]
  \centering
  \small
  \setlength{\tabcolsep}{4pt}
  \resizebox{\columnwidth}{!}{%
  \begin{tabular}{l cc cc cc}
    \toprule
    & \multicolumn{2}{c}{flame steak} & \multicolumn{2}{c}{painter} & \multicolumn{2}{c}{big kicks} \\
    \cmidrule(lr){2-3}\cmidrule(lr){4-5}\cmidrule(lr){6-7}
    Device & $\rho=1$ & $\rho=0.35$ & $\rho=1$ & $\rho=0.35$ & $\rho=1$ & $\rho=0.35$ \\
    \midrule
    \multicolumn{7}{l}{\emph{Decoding a GOP} (ms)$\downarrow$} \\
    iPhone 17       & 167 & 55 & 182 & 68 & 324 & 111 \\
    MacBook Air M2  & 103 & 38 & 130 & 51 & 243 & 79 \\
    \midrule
    \multicolumn{7}{l}{\emph{Playback} (fps)$\uparrow$} \\
    iPhone 17       & 43 & 64 & 34 & 61 & 45 & 76 \\
    MacBook Air M2  & 67 & 108 & 56 & 111 & 79 & 148 \\
    \bottomrule
  \end{tabular}}
  \caption{\textbf{Decoding and playback in the browser}, Chrome. Each GOP is decoded five times after a warm-up, and we report the median over the five GOPs of each scene. Playback decodes and renders all 300 frames in order at the held-out view, and we report the median of three passes. Both render at the full resolution.}
  \label{tab:browser}
\end{table}

%% file: sec/tab_beta.tex
\begin{table}[h]
  \centering
  \small
  \setlength{\tabcolsep}{4pt}
  \resizebox{\columnwidth}{!}{%
  \begin{tabular}{ll ccccc}
    \toprule
    Scene & $\rho=0.35$ & MB/fr$\downarrow$ & PSNR$\uparrow$ & LPIPS$\downarrow$ & Moving PSNR$\uparrow$ & Worst 10\%$\downarrow$ \\
    \midrule
    \multirow{5}{*}{flame steak} & random              & 0.061 & 33.98 & 0.073 & 27.38 & 0.36 \\
                                 & $\beta=1$           & 0.061 & 33.95 & 0.072 & 26.94 & 0.44 \\
                                 & $\beta=8$           & 0.061 & 34.04 & 0.069 & 27.90 & 0.30 \\
                                 & $\beta=32$ (ours)   & 0.061 & 33.92 & 0.071 & 27.96 & 0.40 \\
                                 & $\beta=128$         & 0.060 & 33.72 & 0.074 & 28.02 & 0.75 \\
    \midrule
    \multirow{5}{*}{painter}     & random              & 0.080 & 34.77 & 0.057 & 33.27 & 2.35 \\
                                 & $\beta=1$           & 0.084 & 36.69 & 0.047 & 34.37 & 1.23 \\
                                 & $\beta=8$           & 0.084 & 36.19 & 0.048 & 34.53 & 1.39 \\
                                 & $\beta=32$ (ours)   & 0.084 & 36.46 & 0.048 & 34.65 & 0.98 \\
                                 & $\beta=128$         & 0.083 & 36.12 & 0.052 & 34.63 & 1.24 \\
    \midrule
    \multirow{5}{*}{big kicks}   & random              & 0.165 & 25.56 & 0.149 & 21.58 & 0.39 \\
                                 & $\beta=1$           & 0.161 & 25.73 & 0.135 & 21.69 & 0.12 \\
                                 & $\beta=8$           & 0.161 & 25.72 & 0.136 & 21.67 & 0.12 \\
                                 & $\beta=32$ (ours)   & 0.160 & 25.74 & 0.137 & 21.68 & 0.13 \\
                                 & $\beta=128$         & 0.161 & 25.74 & 0.137 & 21.68 & 0.13 \\
    \midrule
    \multirow{5}{*}{mean}  & random            & 0.102 & 31.44 & 0.093 & 27.41 & 1.03 \\
                                 & $\beta=1$         & 0.102 & 32.12 & 0.085 & 27.67 & 0.60 \\
                                 & $\beta=8$         & 0.102 & 31.98 & 0.084 & 28.03 & 0.60 \\
                                 & $\beta=32$ (ours) & 0.102 & 32.04 & 0.085 & 28.10 & 0.50 \\
                                 & $\beta=128$       & 0.101 & 31.86 & 0.088 & 28.11 & 0.71 \\
    \bottomrule
  \end{tabular}}
  \caption{\textbf{Effect of the motion weight $\beta$} in the pruning score of \cref{eq:prune_motion}, on the first GOP of each scene after adaptation, at the same keep fraction. Moving PSNR covers the pixels whose color changes from the previous frame. Worst 10\% is the mean PSNR loss of the six worst of the 60 frames against the unpruned result. On big kicks, PSNR uses the person mask and LPIPS its bounding box. Mean averages the three scenes.}
  \label{tab:beta}
\end{table}

%% file: sec/tab_lossy.tex
\begin{table}[h]
  \centering
  \small
  \setlength{\tabcolsep}{4pt}
  \resizebox{\columnwidth}{!}{%
  \begin{tabular}{l cc cc cc}
    \toprule
    & \multicolumn{2}{c}{$\rho=1$} & \multicolumn{2}{c}{$\rho=0.5$} & \multicolumn{2}{c}{$\rho=0.35$} \\
    \cmidrule(lr){2-3}\cmidrule(lr){4-5}\cmidrule(lr){6-7}
    Dataset & Saved & $\Delta$PSNR & Saved & $\Delta$PSNR & Saved & $\Delta$PSNR \\
    \midrule
    N3DV        & 14.5\% & $-0.03$ & 14.7\% & $-0.08$ & 14.7\% & $-0.10$ \\
    Technicolor & 14.2\% & $+0.01$ & 14.4\% & $+0.21$ & 14.1\% & $-0.06$ \\
    DanceNet3D  & 14.7\% & $-0.03$ & 14.2\% & $-0.03$ & 14.2\% & $-0.02$ \\
    \midrule
    & \multicolumn{2}{c}{flame steak} & \multicolumn{2}{c}{painter} & \multicolumn{2}{c}{big kicks} \\
    \cmidrule(lr){2-3}\cmidrule(lr){4-5}\cmidrule(lr){6-7}
    Lossy images, $\rho=0.35$ & Saved & $\Delta$PSNR & Saved & $\Delta$PSNR & Saved & $\Delta$PSNR \\
    \midrule
    Rotations                          & 6.4\%  & $-0.23$ & 7.0\%  & $-0.01$ & 7.0\%  & $0.00$ \\
    Center, low byte                   & 8.3\%  & $-0.19$ & 7.9\%  & $+0.20$ & 8.0\%  & $+0.01$ \\
    Scales                             & 4.5\%  & $-0.25$ & 5.6\%  & $+0.51$ & 5.0\%  & $0.00$ \\
    Opacity                            & 1.3\%  & $-0.18$ & 1.2\%  & $-0.14$ & 2.0\%  & $+0.01$ \\
    DC color                           & 3.4\%  & $-0.22$ & 2.6\%  & $-0.28$ & 2.0\%  & $+0.04$ \\
    Rotations, scales                  & 11.5\% & $-0.12$ & 12.6\% & $-0.75$ & 11.6\% & $-0.02$ \\
    Rotations, center low byte         & 14.9\% & $-0.19$ & 14.8\% & $+0.12$ & 14.4\% & $0.00$ \\
    Rotations, scales, center low byte & 19.6\% & $-0.21$ & 20.5\% & $-0.15$ & 19.0\% & $-0.03$ \\
    \quad with low byte at quality 85 & 16.8\% & $-0.18$ & 16.6\% & $+0.21$ & 16.2\% & $0.00$ \\
    \quad with low byte at quality 75 & 19.5\% & $-0.12$ & 19.0\% & $+0.17$ & 18.4\% & $0.00$ \\
    \bottomrule
  \end{tabular}}
  \caption{\textbf{Lossy coding of attribute images.} Saved is the share of bytes saved against our lossless encoder with the same adaptation, and $\Delta$PSNR is the change in dB against it, with PSNR$_\text{fg}$ on DanceNet3D. Top: lossy rotations and center low byte on the first GOP of every scene, averaged per dataset. Bottom: other choices of lossy images on the first GOP of three scenes, at quality 95, and 90 for the low byte of the centers, unless stated.}
  \label{tab:supp_lossy}
\end{table}

%% file: sec/tab_residual.tex
\begin{table}[h]
  \centering
  \small
  \setlength{\tabcolsep}{3pt}
  \resizebox{\columnwidth}{!}{%
  \begin{tabular}{l c c c c c c c cc}
    \toprule
    & & & & Decode & Peak & First & Records & \multicolumn{2}{c}{Lossy $\Delta$PSNR} \\
    \cmidrule(lr){9-10}
    Scene & $\rho$ & Bytes & $\Delta$PSNR & time & memory & frame & damaged & Abs. & Res. \\
    \midrule
    flame steak & 1    & $-22.2$\% & $+0.10$ & 1.12$\times$ & 2.29$\times$ & 0.53$\times$ & 18.5 & -- & -- \\
                & 0.35 & $-18.4$\% & $+0.02$ & 1.19$\times$ & 2.27$\times$ & 0.60$\times$ & 13.9 & $-1.08$ & $-3.06$ \\
    painter     & 1    & $-16.9$\% & $+0.61$ & 0.97$\times$ & 2.22$\times$ & 0.51$\times$ & 8.7 & -- & -- \\
                & 0.35 & $-13.8$\% & $-0.02$ & 1.24$\times$ & 2.14$\times$ & 0.72$\times$ & 7.7 & $-1.27$ & $-4.44$ \\
    big kicks   & 1    & $-21.0$\% & $+0.02$ & 1.23$\times$ & 2.42$\times$ & 0.37$\times$ & 24.6 & -- & -- \\
                & 0.35 & $-15.4$\% & $-0.02$ & 1.19$\times$ & 2.42$\times$ & 0.34$\times$ & 15.8 & $+0.11$ & $+0.34$ \\
    \bottomrule
  \end{tabular}}
  \caption{\textbf{Versions stored as residuals of their predecessors} against our absolute versions, on the first GOP. Bytes, $\Delta$PSNR, decode time per frame, peak GPU memory during decoding and time to the first frame give the residual form relative to the absolute form. Records damaged is the mean number of records that one corrupted record makes wrong, which is 1 for absolute versions. Lossy $\Delta$PSNR is the change in dB when the rotation grids and the low center byte are coded with lossy WebP at quality 90 without adaptation.}
  \label{tab:supp_residual}
\end{table}